\documentclass[10pt,journal,compsoc]{IEEEtran}
\usepackage[utf8]{inputenc}
\usepackage{ntheorem}
\usepackage{graphicx}
\usepackage{comment}
\usepackage{amsmath}
\usepackage[breaklinks=true]{hyperref}
\usepackage{pdflscape}
\usepackage{pifont}
\newcommand{\cmark}{\ding{51}}%
\newtheorem{defn}{Definition}

\title{A Survey of Single-Scene\\Video Anomaly Detection}
\author{anonymous authors}
\author{Bharathkumar~Ramachandra,~\IEEEmembership{Member,~IEEE,}
        Michael~J.~Jones,~\IEEEmembership{Senior Member,~IEEE,}
        and~Ranga~Raju~Vatsavai,~\IEEEmembership{Member,~IEEE}
\IEEEcompsocitemizethanks{\IEEEcompsocthanksitem B. Ramachandra and R. Vatsavai are with the Department
of Computer Science, North Carolina State University, Raleigh,
NC 27695.\protect\\
E-mail: bramach2@ncsu.edu, rrvatsav@ncsu.edu
\IEEEcompsocthanksitem M. Jones is with Mitsubishi Electric Research Labs (MERL), Cambridge, MA 02139.\protect\\
Email: mjones@merl.com}
}

\begin{document}

\IEEEtitleabstractindextext{%
\begin{abstract}
This survey article summarizes research trends on the topic of anomaly detection in video feeds of a single scene. We discuss the various problem formulations, publicly available datasets and evaluation criteria. We categorize and situate past research into an intuitive taxonomy and provide a comprehensive comparison of the accuracy of many algorithms on standard test sets. Finally, we also provide best practices and suggest some possible directions for future research.
\end{abstract}

}

\maketitle

\IEEEdisplaynontitleabstractindextext

\IEEEpeerreviewmaketitle


\section{Introduction}
\label{sect:introduction}
\IEEEPARstart{V}{ideo} anomaly detection is the task of localizing anomalies in space and/or time in a video, where anomalies are simply activities that are out of the ordinary. Anomalies are also referred to as abnormalities, novelties, and outliers among other similar terms. Examples range from unattended bags at airports, to people falling down, to a person loitering outside a building. We follow the definition provided in \cite{saligrama_video_2010},
\begin{defn}
Video anomalies can be thought of as the occurrence of unusual appearance or motion attributes or the occurrence of usual appearance or motion attributes in unusual locations or times.
\label{defn:video-anomalies}
\end{defn}

One implication of this definition that is not immediately obvious is that video anomalies are \textit{scene-dependent}. This means that activity that is anomalous in one scene may be normal in another. For example, in one scene, riding a bicycle along a bike path is normal, while in another, riding a bicycle down a similar looking pedestrian sidewalk is anomalous. Normal video (that is, video not containing any anomalies) is thus needed for model training to express the variety of normal activities that may occur in a particular scene. Since it is unrealistic to collect video for all possible anomalous events for training and expensive to collect even a few anomalous events, a common assumption is that training data consists of \textit{only} normal activities which is relatively easy to obtain.

This survey focuses on single-scene video anomaly detection because it is the most common use case for video anomaly detection in real-world applications.  The motivating example is a surveillance camera monitoring a scene and a person responsible for noticing any unusual activity that occurs. This scenario highlights the practical importance of developing algorithms for single-scene video anomaly detection, because this is clearly a task that would be better done by a computer given the extreme difficulty for a person to pay attention to a camera feed (typically with nothing interesting occurring) for long periods of time. If this scenario were changed to a person monitoring a bank of camera feeds, it would still be best modeled as several single-scene video anomaly detection problems for two reasons: (1) the need to handle location-dependent anomalies and (2) the possibility that all the scenes in the camera feeds are not consistent.

Most prior work on video anomaly detection has not recognized the important distinction between single-scene video anomaly detection and multi-scene.  One important difference is that the single-scene anomaly detection formulation can contain location-dependent anomalies whereas multi-scene cannot. The lack of recognition of the single-scene/multi-scene distinction is likely due to the fact that most single-scene video anomaly detection datasets (Street Scene being the main exception) contain very few location-dependent anomalies which means that methods that do not accommodate location-dependent anomalies are not heavily penalized. A location-dependent anomaly is an object or activity that is anomalous in some regions of a scene but not in other regions.  A good example is walking on grass. In a particular scene, there may be some areas of grass that are normal to walk on and other areas that are restricted and thus anomalous to walk on. The only factor that distinguishes these two activities is the location. In multi-scene video anomaly detection, normal video from many different, unrelated scenes are given for building a single model of normality.  The goal in this case is to learn the normality in a variety of appearances and activities that occur \textit{anywhere in any of the videos}.  Because there is no correspondence across scenes in the multi-scene formulation it is not possible to create a single model in which an activity is anomalous in some locations of some scenes but not in other locations.  Location-dependent anomalies (such as jaywalking, riding a bike on a pedestrian sidewalk, driving a car in the wrong direction and etcetera) which involve normal activity or objects occurring in unusual locations are commonplace for single-scene video anomaly detection.

Another important consideration for multi-scene video anomaly detection that does not apply to single-scene is that the normal training videos need to be ``consistent'' in the sense that what is normal and what is anomalous must be the same in all of the scenes.  This is because a single model of normality is being built from video of all the different scenes.  For example, a truck backing up to a building may be normal in one scene because there is a loading dock while a truck backing up to a building in another scene may be anomalous.  Such scenes would not be consistent.

Because of the practical importance of single-scene video anomaly detection, we focus on this formulation of the problem in this survey.

The currently available datasets for single-scene video anomaly detection (UCSD Ped1 \& Ped2 \cite{weixin_li_anomaly_2014}, CUHK Avenue \cite{lu_abnormal_2013}, Subway \cite{adam_robust_2008}, UMN \cite{UMN}, and Street Scene \cite{ramachandra_street_2019}) are also static-camera datasets, but the camera being static is not necessary.  In fact, the CUHK Avenue \cite{lu_abnormal_2013} and Street Scene \cite{ramachandra_street_2019} datasets do have some minor camera motion. One can imagine a model for a single scene being able to handle camera motion when the majority of each frame overlaps with neighboring frames (as would occur with a pan-tilt-zoom surveillance camera) by keeping track of global location in each frame. Such a formulation would still be considered single-scene. There are currently no benchmark datasets available or algorithms proposed for this single-scene moving-camera version of the problem, but it is a fertile area for future research.

\begin{figure*}
    \centering
    \includegraphics[width=0.9\linewidth]{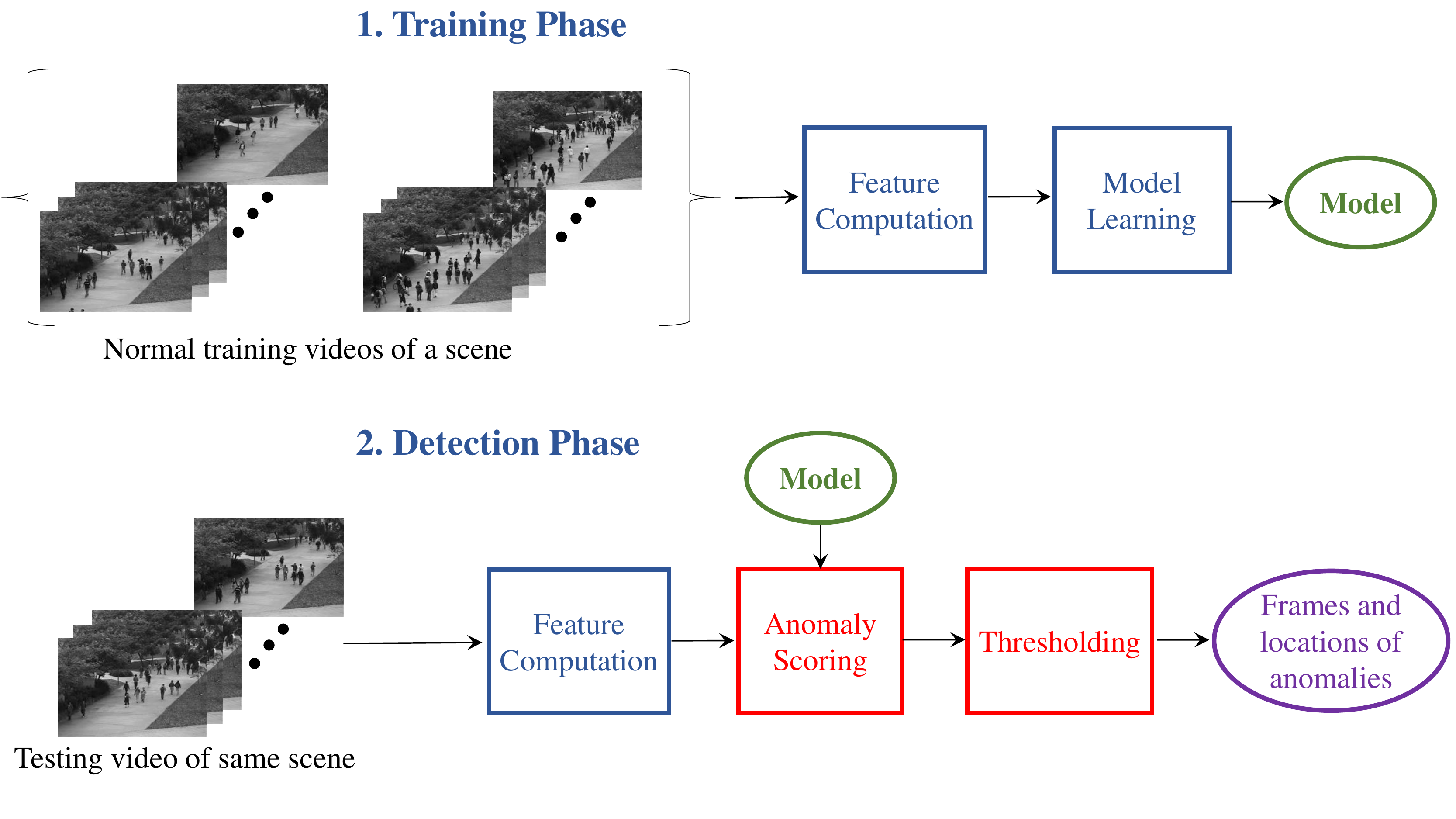}
    \vspace{-10pt}
    \caption{Overview of single-scene video anomaly detection.  Typical algorithms include a model building phase in which a model of normal activity is learned from one or more videos of a scene followed by a detection phase in which anomalies are detected in video from the same scene.}
    \label{fig:overview}
    \vspace{-5pt}
\end{figure*}

Figure \ref{fig:overview} shows an overview of typical algorithms for single-scene video anomaly detection.  First, during a training phase, a model of normal activity is learned from the features computed from one or more videos of a scene which do not contain anomalies.  Then in the detection phase, new video from the same scene is given from which the same types of features are computed.  The features along with the model are used to assign anomaly scores to each voxel of the input video. Anomaly scores are then thresholded to yield  spatio-temporal binary masks of the anomalies detected.

\subsection{Other formulations of the problem}
\label{subsect:other-formulations}
It is important to point out that many papers on video anomaly detection have addressed different formulations of the problem than the single-scene formulation on which this survey focuses.  We have already discussed the multi-view formulation (\cite{morais2019learning,sultani_real-world_2018,liu_future_2018,hasan_learning_2016,ionescu_object-centric_2019}) above in some detail.  

Another alternate formulation for video anomaly detection that has been used in a number of papers (\cite{leibe_discriminative_2016,zhao_online_2011,ionescu_unmasking_2017,liu2018classifier,pang2020self}) is \textit{training-free} video anomaly detection.  In this formulation, no normal training video is provided and the task is to either detect changes in the testing video or else to detect the most unusual segments of the testing video as proxies for anomalousness.  Detecting the most unusual segments of testing video is analogous to discord detection in time series analysis \cite{KeoghEtAl2005}.  While these formulations of the problem are also useful, they are significantly different from single-scene video anomaly detection and require different datasets and ground truth annotations.

Many existing research papers do not clearly distinguish which problem formulation they are using.  This leads to ambiguities and confusion about what datasets should be tested on and which methods should be compared against. It also leads to differences in understanding the performance of different methods. We think it is important to make clear the problem formulation being used in any paper on video anomaly detection. In this survey, we have chosen to focus on the single-scene video anomaly detection formulation because it encompasses a very common scenario and has many practical applications, such as in surveillance, security, factory automation (monitoring the activity of workshop floors), video search and video summarization.

\subsection{Types of Video Anomalies}
\label{subsect:anomaly-types}
Here we attempt to provide a non-exhaustive list of what we think are the most commonly occurring video anomalies; a specific application may warrant the declaration of other types of anomalies.

\subsubsection{Appearance-only Anomalies}
\label{subsubsect:app-only}
These anomalies can be thought of as unusual object appearance in a scene.  Examples include bicyclists on a pedestrian walkway, or a large boulder on a road.  Detecting these anomalies only requires inspecting a local region of a single frame of video.

\subsubsection{Short-term Motion-only Anomalies}
\label{subsubsect:short-term-motion}
These anomalies can be thought of as unusual object motion in a scene.
Examples include a person running in a library, or a car skidding sideways on the road. Detecting these anomalies usually only require inspecting a local region of the video over a short period of time.
Appearance-only and short-term motion-only anomalies can be further called \textbf{local} anomalies because they possess this additional property.

\subsubsection{Long-term Trajectory Anomalies}
\label{subsubsect:traj}
These anomalies can be thought of as unusual object trajectory in a scene.
Examples include persons walking in a zig-zag fashion on a sidewalk, a car weaving in and out of traffic, or loitering around foreign embassy buildings. Detecting trajectory anomalies requires inspecting longer segments of video.

\subsubsection{Group Anomalies}
\label{subsubsect:group}
Group anomalies can be thought of as unusual object interaction in a scene. An example is a group of persons walking in a formation (such as a marching band). Detecting group anomalies requires analyzing the relationship between two or more regions of video.

\subsubsection{Time-of-Day Anomalies}
\label{subsubsect:time-of-day}
This type of anomaly is orthogonal to all of the other types. What makes these activities anomalous is \textit{when} they happen. These anomalies are in spirit very similar to the location-dependent anomalies discussed earlier, with the ``relevant contextual frame of reference'' being temporal instead of spatial.
An example is when persons enter a movie theatre at the early hours of dawn.  
Usually, detecting these anomalies just requires using a different model of normality for different times of day.

\hfill \break
\textbf{A note on the types of anomalies}

Not all of these different types of anomalies may be necessary to detect for every application. Thus, video anomaly detection is further, \textit{application dependent}. In fact, \textit{in the publicly available datasets for video anomaly detection that we describe, mainly only appearance-only and short-term motion-only anomalies occur}. 
We should also note that the different types of anomalies are not mutually exclusive. In fact, it can be hard to come up with examples that are exclusive to some of the types listed above.

\hfill \break
\textbf{Anomalousness is a continuous measure}

It is important to note that although often discussed in the binary sense, anomalousness is a fluid concept. Every activity is anomalous \textit{to some extent}.  For example, in a scene of a pedestrian walkway, a tall man in a red shirt walking at 1 meter/second may not have been seen exactly in the normal video, but he is most likely similar to some pedestrians in the normal video and therefore should be assigned a low anomaly score.  However, if the man is 3 meters tall or walking at 10 meters/second then this should presumably receive a much higher anomaly score.  Finding features and distance measures that correspond to our intuitive notions of when two activities are similar is a key to creating successful video anomaly detection algorithms.

\subsection{Other Considerations for Video Anomaly Detection}
\label{subsect:other-considerations}
Here we discuss some other characteristics of video anomaly detection formulations that vary in past work on the topic.

\subsubsection{Unsupervised, Semi-supervised, Weakly Supervised or Supervised?}
\label{subsubsect:supervision}
The anomaly detection problem is difficult to neatly characterize. Should it be called unsupervised because examples of anomalies have not been provided for supervision? Or should it be called supervised because normal data is provided for supervision? Or how about semi-supervised because only selective data (normal) is provided for training? Some call this problem weakly-supervised because an auxiliary dataset is necessary to determine (provide supervision for) an anomaly score threshold or because proxy labels are often used. We discuss another possible formulation that some have considered in Section \ref{subsubsect:video-level-weak-sup}. We assert that summarizing the formulation with these terms is suboptimal and causes confusion for readers. We recommend that future video anomaly detection research works should always provide a full description of the problem formulation considered to avoid any ambiguities and new methods compare against methods that follow compatible formulations.

\subsubsection{Temporal Localization or Spatial Localization Too?}
\label{subsubsect:localization}
Although several past works have focused solely on the temporal (frame-level) localization aspect of this problem (\cite{sultani_real-world_2018,liu_future_2018,hasan_learning_2016,chong_abnormal_2017}), we contend that spatial localization is paramount to a useful algorithm. Solely temporal localization is useful for very limited applications such as key-frame prediction and video compression, but even in these cases it is useful to know which parts of the frame were deemed anomalous. In general, in a busy scene, knowing only that \textit{something} in the frame is anomalous may leave the user wondering exactly what triggered the anomaly detector.  We clearly outline which past works focus solely on temporal localization and which include spatial localization as well.

\subsection{Other Surveys}
\label{subsect:other-surveys}
Two past surveys focus on crowded scene analysis (\cite{zhan_crowd_2008,li_crowded_2015}), which is important and relevant to successful video anomaly detection, but these surveys are not primarily concerned with video anomaly detection.  A survey by Sodemann et al. \cite{sodemann_review_2012} focused on anomaly detection in surveillance videos, but is a high-level view of the area, does not cover the most recent work and does not include a comprehensive performance evaluation of different algorithms as our survey does.  A short survey by Chong et al.~\cite{chong_modeling_2015} from 2015 is narrowly focused on different methods of modeling video and does not include a comparison of methods on video anomaly detection datasets.  Finally, a survey by Kiran et al.~\cite{kiran_overview_2018} from 2018 focuses mainly on reconstruction approaches to video anomaly detection and also does not provide a comprehensive comparison across many methods in the field.  Unlike past surveys, ours includes a discussion and categorization of a broad selection of methods for video anomaly detection, a quantitative comparison of many different algorithms on standard datasets, a discussion of the important publicly available datasets, a discussion of various evaluation criteria, as well as recent trends and directions for future research.

The rest of this article is organized as follows. In Section \ref{sect:overview}, we describe the publicly available benchmark datasets along with the evaluation protocol for video anomaly detection and set up a taxonomy for the rest of the paper. In Sections \ref{sect:dist-based}, \ref{sect:stats} and \ref{sect:recon-based} we describe notable past works which have employed different approaches to video anomaly detection. In Section \ref{sect:comp-study} we present a comparative study between the various methods. Finally, in Section \ref{sect:discussion} we discuss the state of research in this field and provide some recommendations for future research directions.

\section{Overview of single-scene video anomaly detection}
\label{sect:overview}
\subsection{Datasets}
\label{subsect:datasets}
\begin{figure*}
    \centering
    \includegraphics[width=\linewidth]{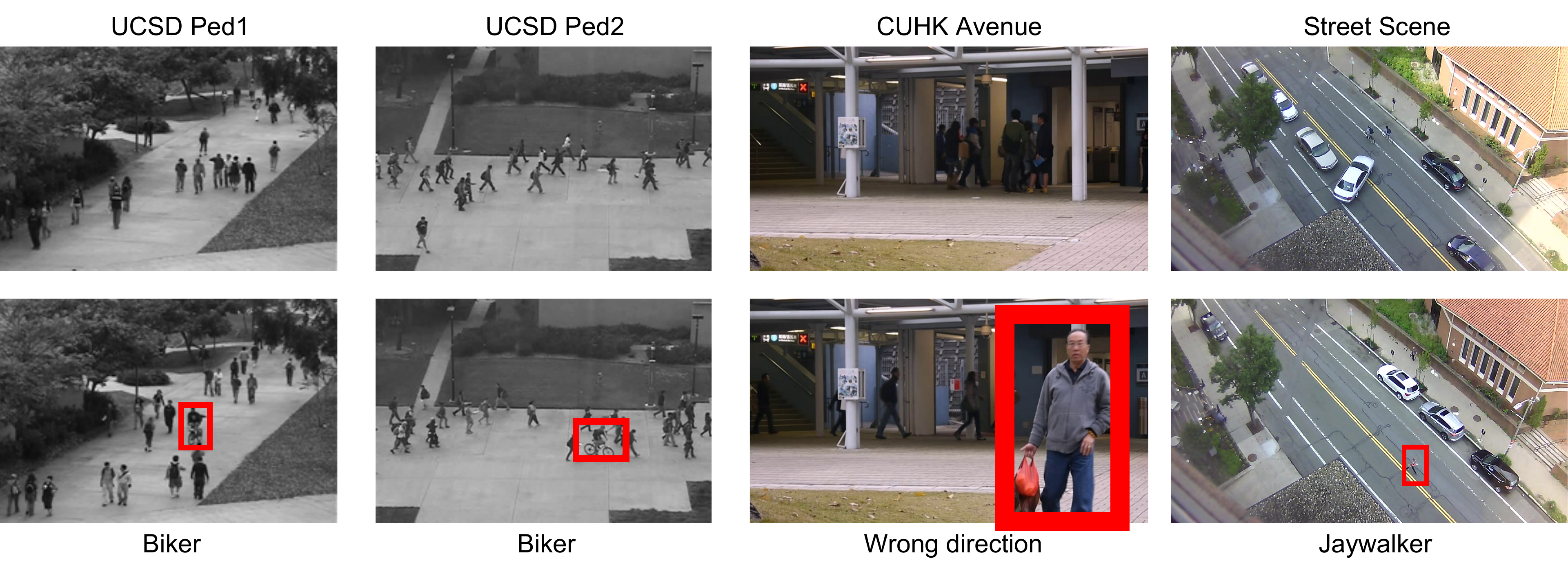}
    \vspace{-10pt}
    \caption{One normal frame and one frame with an anomaly from each of the recommended datasets for single-scene video anomaly detection.}
    \vspace{-5pt}
    \label{fig:datasets}
\end{figure*}

Benchmark datasets play an important role in the progress of research for any problem in computer vision. They help to define the scope of the problem as well as provide a way to fairly compare the characteristics of different algorithms. For video anomaly detection, there are a handful of publicly available benchmark datasets in common use.  We describe them here and provide recommendations based on ground-truth annotation style, size and overall utility of the datasets. Table \ref{tab:datasets} provides a summary of the characteristics of these datasets and Figure \ref{fig:datasets} shows one normal frame and one frame with a single anomaly for each of the datasets we recommend for use.

\begin{table}[ht]
\caption{Characteristics of Video Anomaly Detection Datasets. $^*$Aggregates from 2 Scenes.  $^{**}$Aggregates from 3 Scenes. Adapted from \cite{ramachandra_street_2019}.}
\vspace{-15pt}
\label{tab:datasets}
\begin{center}
\resizebox{\linewidth}{!}{
\begin{tabular}{|c|r|r|r|r|c|c|} 
\hline
\textbf{Dataset}                            & \textbf{Total}                             & \textbf{Training}                    & \textbf{Testing}                      & \textbf{Anomalous}              & \textbf{Pixel-wise}   & \textbf{Track ID}  \\
                                   & \textbf{Frames}                            & \textbf{Frames}                      & \textbf{Frames}                       & \textbf{Events}                  & \textbf{annotation} & \textbf{annotation}       \\ 
\hline
Subway$^*$   & 125,475      & 22,500 & 102,975 & 85 & N            & N                \\ 
\hline
UMN$^{**}$  & 3,855 & N/A    & N/A     & 11 & N            & N                \\ 
\hline \hline
 UCSD Ped1, Ped2$^*$            & 18,560                            & 9,350                       & 9,210                        & 77                      & Y            & Y                \\ 
\hline
CUHK Avenue                        & 30,652                            & 15,328                      & 15,324                       & 47                      & N \footnotemark           & Y                \\ 
\hline
Street Scene                       & 203,257                           & 56,847                      & 146,410                      & 205                     & N            & Y                \\
\hline
\end{tabular}
}
\end{center}
\end{table}
\footnotetext{Note that although the authors' original annotations are at the pixel-level, the pixel masks are all from bounding boxes, not of object silhouettes as is the case with the UCSD datasets.} 

\subsubsection{Subway}
The Subway dataset \cite{adam_robust_2008} is comprised of two long videos of two different indoor scenes, a subway entrance and an exit, making for 2 separate datasets. It mainly captures people entering or leaving through turnstiles. Anomalies include people jumping or squeezing through turnstiles, a janitor cleaning the walls and people walking in the wrong direction. It is unclear at what frame rate one should extract the dataset from these videos and exactly which frames are labeled as anomalous and which frames to use for training and testing. Table \ref{tab:datasets} uses 15 frames/sec to obtain the frame counts.  No spatial ground truth is provided. The datasets contain 85 total anomalous events labeled temporally. These datasets are now quite old and because of the ambiguities and lack of spatial annotation, we do not recommend using these for evaluating an anomaly detection method in any formal capacity. Those seeking the datasets should contact the author directly.

\subsubsection{UMN}
\label{subsubsect:UMN}
The UMN dataset~\cite{UMN} has 11 short clips from 3 different cameras at an outdoor field, an outdoor courtyard and an indoor foyer. All clips start with normal activity followed by an anomalous event where the crowd suddenly disperses quickly, hinting at an evacuation scenario. The anomalies are staged and every clip has exactly one anomalous event. There is no clear specification about frame rate for extraction or a training or test split. Fifteen frames/sec was used for the frame counts in Table \ref{tab:datasets}.  Additionally, anomalies are only labeled temporally. Because of these ambiguities and the lack of spatial annotation, we do not recommend using it for evaluating an anomaly detection method in any formal capacity.

The dataset and ground truth can be found at \url{http://mha.cs.umn.edu/proj_events.shtml#crowd}.

\subsubsection{UCSD Pedestrian}
\label{subsubsect:UCSDPed}
The most widely used datasets for video anomaly detection are the UCSD Ped1 and Ped2 datasets \cite{mahadevan_anomaly_2010,weixin_li_anomaly_2014}. Each of these datasets contains videos from a different static camera overlooking a pedestrian walkway, and the crowd density is sometimes high to the point of causing severe occlusions.  In this dataset, all non-pedestrian objects as well as unusual motion by pedestrians are deemed anomalous. The types of anomalies present are ``biker'', ``skater'', ``cart'', ``wheelchair'', ``walk across'', and ``other''. UCSD Ped1 consists of 34 training videos and 36 testing videos at a low spatial resolution of $158 \times 238$ pixels. The field-of-view can be considered mid-range and there are 200 frames per video. UCSD Ped2 contains 16 training and 12 testing videos of slightly higher resolution, $240 \times 360$ pixels, with 120 to 200 frames per video. 

These datasets can be found at \url{http://www.svcl.ucsd.edu/projects/anomaly/dataset.htm}. Both spatial (at the pixel-level) and temporal annotation are available for UCSD Ped1 and Ped2 datasets from the authors. One should note that the authors only released partial pixel-wise ground truth for UCSD Ped1, which was subsequently completed by the authors of \cite{antic_video_2011} and made available at \url{https://hci.iwr.uni-heidelberg.de/COMPVIS/research/parsing/}. Very recently, the authors in \cite{vu_robust_2019} released their ``corrected'' set of pixel-level annotations as well, claiming that the original annotation has errors at \url{https://github.com/SeaOtter/vad_gan}. Another set of bounding box annotations containing anomalous region identifiers as well as track identifiers required for evaluating using a more recent criteria has been made available by the authors of \cite{ramachandra_learning_2019} at \url{http://www.merl.com/demos/video-anomaly-detection}. 

\subsubsection{CUHK Avenue}
\label{subsubsect:avenue}
The CUHK Avenue dataset \cite{lu_abnormal_2013} consists of short video clips taken from a single camera looking at the side of a building with pedestrian walkways by it. The videos mainly contain people walking in and out of a building. Concrete columns that are part of the building cause some severe occlusion. In \cite{lu2019fast}, the authors double the size of the dataset and label spatial locations of the abnormal events. The dataset contains 16 training videos and 21 testing videos of spatial resolution $640 \times 360$ pixels. There are a total of 47 anomalous events which are mostly staged and comprise actions such as ``throwing papers'', ``throwing bag'', ``child skipping'', ``wrong direction'' and ``bag on grass''. We should note that this was the first dataset to introduce a loitering (static) anomaly with the bag, which is important for surveillance applications.

Both temporal and pixel-level (in bounding box form) annotations are provided by the authors. The dataset and ground truth can be found at \url{http://www.cse.cuhk.edu.hk/leojia/projects/detectabnormal/dataset.html}. Another set of bounding box annotations containing anomalous region identifiers as well as track identifiers required for evaluating using more recent protocol has been made available by the authors of \cite{ramachandra_learning_2019} at \url{http://www.merl.com/demos/video-anomaly-detection}. 

Researchers should be aware that some papers that report results on Avenue used some evaluation code available on GitHub (\url{https://alliedel.github.io/anomalydetection/}) that incorrectly computes pixel-level results \cite{leibe_discriminative_2016,ionescu_unmasking_2017,battiato_deep_2017,ionescu_detecting_2019}.  The code produced pixel-level area under the curve (AUC) numbers that were higher than frame-level AUC numbers, which is not possible since frame-level AUC imposes an upper bound on pixel-level AUC.  Future papers should not cite these incorrect results and should not use the buggy code that produced them.

\subsubsection{Street Scene}
\label{subsubsect:streetscene}
The largest dataset, Street Scene \cite{ramachandra_street_2019}, is the most recent addition to the publicly available datasets for video anomaly detection. Street Scene consists of 46 training and 35 testing high resolution $1280 \times 720$ video sequences taken from a USB camera overlooking a scene of a two-lane street with bike lanes and pedestrian sidewalks during daytime. The dataset is challenging because of the variety of activity taking place such as cars driving, turning, stopping and parking; pedestrians walking, jogging and pushing strollers; and bikers riding in bike lanes. In addition the videos contain changing shadows, moving background such as a flag and trees blowing in the wind, and occlusions caused by trees and large vehicles. There are a total of 56,847 frames for training and 146,410 frames for testing, extracted from the original videos at 15 frames per second. The dataset contains a total of 205 naturally occurring anomalous events ranging from illegal activities such as jaywalking and illegal U-turns to simply those that do not occur in the training set such as pets being walked and a metermaid ticketing a car. We refer readers to \cite{ramachandra_street_2019} for a more detailed description with complete meta-data.

The authors make the dataset available along with a set of bounding box annotations containing anomalous region identifiers as well as track identifiers required for evaluating on more recent protocols (that they also introduced) at \url{http://www.merl.com/demos/video-anomaly-detection}.

\subsubsection{Other Datasets}
\label{subsubsect:other-datasets}
It is worth noting a few other datasets that are useful for multi-scene video anomaly detection.  Because these datasets include videos from various unconnected scenes and from which a single model is meant to be learned, they are not applicable to the single-scene video anomaly detection formulation that this survey focuses on.

\textbf{ShanghaiTech}

ShanghaiTech ~\cite{liu_future_2018} is a recent contribution that contains videos from 13 different scenes. A typical video has people walking along a sidewalk of a university. Anomalous activity includes bikers, skateboarders and people fighting.  The dataset is intended to be used to learn a single model from the training sets of all 13 scenes.  While it is conceivable to treat this dataset as 13 separate datasets (as with UCSD Ped 1 and Ped2), this is problematic since this would yield an average of 10 anomalous events per scene which is very small, and it is not clear whether the variation captured in each scene's small training set is meant to serve as representative of normal activity. The dataset is available for download at \url{https://svip-lab.github.io/dataset/campus_dataset.html}.

\textbf{UCF-Crime}

The UCF-Crime dataset \cite{sultani_real-world_2018} is a recently proposed new dataset for video anomaly detection. This dataset contains 128 hours of internet videos taken from many different cameras and contains criminal anomalous activities such as burglary, shoplifting and assault.  Anomalies are only annotated temporally (i.e. no spatial annotations are available).  The authors also advocate for classifying anomalies according to a predetermined set of anomaly types which makes the problem formulation that this dataset is intended for different from the usual multi-scene video anomaly detection formulation. The dataset can be downloaded from the project page at \url{https://www.crcv.ucf.edu/projects/real-world/}.

\textbf{Car Dashcam Datasets}

Another interesting and large subset of multi-view datasets are dashcam video datasets taken from moving cameras inside of cars and trucks.  These include datasets from \cite{HerzigEtAl2019}, \cite{ChanEtAl2016}, \cite{CheEtAl2019} (called D$^2$-City), and \cite{HareshEtAl2020} (called RetroTrucks).  Anomalies in these datasets mainly consist of traffic accidents.

\subsection{Evaluation Protocol}
\label{subsect:eval-protocol}
It is important to remember that anomalies are scene-dependent and what is anomalous is completely determined by what activity occurs at test time but is missing from the training set (that defines normal activity). Moreover, the ground truth annotations are binary in nature although anomalousness is a fluid notion. Determining which activities are missing from the training video can often lead to ambiguities.  For example, two people walking next to each other along a sidewalk may exist in the training video, but two people holding hands while walking may not.  Should the latter be marked as anomalous?  In which frame exactly does the anomaly begin and end?  Should the entire area including both pedestrians be marked as anomalous or just a tight area around the hand-holding?  Every dataset and annotator for this task is imperfect and ambiguities such as these will exist. Ideally, an evaluation measure would attempt to give a realistic measure of the qualitative performance of an algorithm in practice despite the inevitable ambiguities in labeling.

\subsubsection{Traditional Criteria}
\label{subsubsect:traditional-criteria}
Traditionally, research in this field has used frame-level and pixel-level criteria to evaluate performance, first described in \cite{mahadevan_anomaly_2010} (which also presented the UCSD Pedestrian datasets). These criteria basically describe how to count positives, negatives, true positive and false positives and subsequently compute true positive rate (TPR) and false positive rate (FPR) at a given anomaly score threshold:

$$\mbox{TPR} = \frac{\mbox{num. of true positive frames}}{\mbox{num. of positive frames}}$$

$$\mbox{FPR} = \frac{\mbox{num. of false positive frames}}{\mbox{num. of negative frames}}$$

Then, the threshold on anomaly score is varied in order to generate Receiver Operating Characteristic (ROC) curves of FPR versus TPR. Area under the ROC curve (AUC) and Equal Error Rate (EER) are used to summarize an ROC curve.

These criteria use pixel-level ground truth. That is, every frame at time $t$, $\mathbf{F^t}$, has a corresponding binary ground truth mask $\mathbf{A^t}$ indicating whether or not each pixel is anomalous. 

The frame-level criterion is as follows: Given the predicted per-pixel anomaly score map $\mathbf{S^t}$ corresponding to the $t^{th}$ frame of a test video, the frame is said to be predicted as anomalous if $\sum_p [\mathbf{S^t}(p) \geq \Gamma] \geq 1$ where $p$ indexes over all pixels in a frame and $\Gamma$ is the anomaly score threshold. The notation $[C]$ evaluates to $1$ if condition $C$ is true (or $1$) otherwise it evaluates to $0$.  Further, a frame predicted to be anomalous at time $t$ is counted as a true positive frame if $\sum_p [\mathbf{A^t}(p) == 1] \geq 1$ and as a false positive if $\sum_p [\mathbf{A^t}(p) == 1] == 0$.

In other words, frames are predicted as anomalous if they have at least one pixel that received a score larger than the anomaly score threshold. A frame predicted to be anomalous is counted as a true positive if the annotation for that frame has at least one ground truth anomalous pixel and a false positive otherwise. The total number of positives and negatives are determined by the frame-level annotations and are used to compute TPR and FPR, and subsequently AUC and EER. 

$$\mbox{num. of positive frames} = \sum_{t=1}^{T} (\sum_p[\mathbf{A^t}(p) == 1] \geq 1)$$

$$\mbox{num. of negative frames} = \sum_{t=1}^{T} (\sum_p[\mathbf{A^t}(p) == 1] == 0)$$
where $t$ indexes over testing frames and $T$ is the total number of testing frames.  In other words, the number of positive frames is the number of testing frames with at least one ground truth anomalous pixel while the number of negative frames is the number of testing frames with no ground truth anomalous pixels.

The frame-level criterion does not evaluate whether any spatial localization has been achieved and the authors themselves recommended against using solely this criterion in \cite{weixin_li_anomaly_2014}, instead suggesting use of the pixel-level criterion. 

The pixel-level criterion is as follows: Given the predicted anomaly score map $\mathbf{S^t}$ corresponding to the $t^{th}$ frame of a test video, the frame is counted as a true positive frame if $\sum_p [(\mathbf{S^t}(p) \geq \Gamma) \cdot \mathbf{A^t}(p)] \geq 0.4 \cdot \sum_p [\mathbf{A^t}(p) == 1]$ and $\sum_p [\mathbf{A^t}(p) == 1] \geq 1$. Conversely, the frame is counted as a false positive frame if $\sum_p [\mathbf{S^t}(p) \geq \Gamma] \geq 1$ and $\sum_p [\mathbf{A^t}(p) == 1] == 0$. 

In other words, a frame is counted as a true positive frame if over 40\% of the annotated ground truth anomalous pixels in a frame are predicted as anomalous by the model. If a frame has no ground truth anomalous pixels and yet even one pixel is predicted as anomalous, a false positive is counted. Notice that with this criterion, even though spatial localization is taken into account (albeit crudely), the counting of true positives and false positives is still at the frame level. The total number of positives and negatives are as with the frame-level criterion. This has the following consequences:

\begin{enumerate}
    \item A frame can be counted for only one true positive even if there are multiple anomalies present in the frame. The 40\% threshold is applied over \textit{all ground truth anomalous pixels in a frame}.
    \item A frame that contains at least one ground truth anomalous pixel cannot count as a false positive regardless of any incorrect regions in the frame that are predicted as anomalous.
    \item A frame without any ground truth anomalous pixels can be counted for only one false positive even if there are multiple distinct regions that are predicted as anomalous.
    \item The criterion does not penalize looseness of a predicted region. That is, as long as 40\% of annotated pixels are predicted as anomalous, it does not hurt performance to change the prediction mask to the entire frame.
\end{enumerate}

Notice that as described, frame-level AUC for a method imposes an upper-bound on pixel-level AUC. As the authors in \cite{ramachandra_street_2019} observe, points 2 and 3 above admit a simple post-processing step that makes pixel-level AUC exactly reach its upper bound: dilating prediction masks with a filter of the same size as the frame (i.e. if a single pixel is predicted as anomalous in a frame, make all pixels in the frame anomalous).  At a given threshold, this can \textit{only increase the true positive rate without changing the false positive rate} according to the pixel-level criterion.

We should also note that in \cite{mahadevan_anomaly_2010}, the authors fail to fully describe pixel-level evaluation measure. Specifically, the authors define a true positive as a frame where at least 40\% of the truly anomalous pixels in the frame are predicted as anomalous, and a false positive \textit{otherwise}. In their subsequent work \cite{weixin_li_anomaly_2014}, they clarify that a false positive can only be counted for frames that do not contain any anomaly annotation, that is, a false positive should not be counted when fewer than 40\% of the pixels are predicted as anomalous in a frame that has an anomaly. The clarification makes for a strict reduction in the counts of false positives. We believe that some earlier works might have reported results under the incorrect interpretation of this evaluation metric, leading to much lower pixel-level AUCs being reported.

Although these criteria, if correctly used, can be useful for ranking different video anomaly detection algorithms, they are now saturated on the smaller datasets (frame-level AUCs have repeatedly been reported on the UMN dataset at  $>99\%$ for the past few years) and clearly have serious flaws. 

\subsubsection{Recent Criteria}
\label{subsubsect:recent-criteria}
Several researchers have recognized these drawbacks of the frame-level and pixel-level criteria and a few have attempted to propose new criteria aimed at addressing them.  The authors of \cite{sabokrou_deep-cascade:_2017} proposed the {\em Dual Pixel Level} criterion which adds an additional constraint to the pixel-level criterion. With the same notation as before, a frame at time $t$ would only be counted as a true positive if 
$\sum_p [(\mathbf{S^t}(p) \geq \Gamma) \cdot \mathbf{A^t}(p)] \geq 0.4 \cdot \sum_p [\mathbf{A^t}(p) == 1]$ and $\sum_p [(\mathbf{S^t}(p) \geq \Gamma) \cdot \mathbf{A^t}(p)] \geq 0.1 \cdot \sum_p [\mathbf{S^t}(p) \geq \Gamma]$ and $\sum_p [\mathbf{A^t}(p) == 1] \geq 1$.

That is, in addition to the pixels predicted as anomalous needing to cover at least 40\% of the ground truth anomalous pixels, at least 10\% of the pixels predicted as anomalous also need to be \textit{covered by} the ground truth anomalous pixels. In other words, the pixels predicted as anomalous cannot include too many normal pixels (thus preventing the post-processing filtering mentioned above from helping). While this is an improvement, it still cannot correctly count true positives and false positives in frames with (a) multiple ground truth anomalies, (b) both true positive as well as false positive predicted pixels/regions and (c) multiple false positive predicted pixels/regions.
The authors of \cite{lu2019fast} also realized that the pixel-level criterion is flawed and used object-detection style Intersection Over Union (IOU) to penalize both tightness and looseness of a detection on the CUHK Avenue dataset. Unfortunately, this does not fix the issues with multiple counts of either true positives or false positives. Moreover, they are not able to use this IOU-based criterion on other datasets due to differences in annotation formats.

The authors of \cite{ramachandra_street_2019} proposed two new criteria, region-based and track-based, to replace the previous criteria.  The new criteria are claimed to provide a much more realistic picture of how an algorithm will perform in practice. In their perspective, the evaluation protocol should be designed in such a way as to account for ambiguities, biases and inconsistencies that are to be expected in any anomaly detection dataset. To fix the issues with the old criteria, they essentially take two steps:

\begin{enumerate}
    \item They account for inherent ambiguity in labeling and detecting anomalous events by suggesting a loose object detection style Intersection Over Union (IOU) criterion to judge spatial localization. Further, their track-based criterion only requires that anomalies in a fixed percentage of frames in an anomalous track be detected.
    \item They count true and false positives atomic to a detected region rather than atomic to a frame. This means that under their criteria, a frame can have more than one true or false positive, in line with basic intuition.
\end{enumerate}

These criteria require reasoning about ground truth and detected anomalous regions within frames as opposed to whole frames.  Some annotated datasets specify ground truth as bounding boxes which directly give the ground truth regions. For datasets that specify ground truth as pixelwise masks, a set of anomalous regions can be computed as connected components of anomalous pixels.  Similarly, detected regions can be computed as connected components of detected pixels for algorithms that return pixelwise anomaly masks.

The {\em region-based detection criterion} calculates the region-based detection rate (RBDR) over all frames in the test set versus the number of false positive regions per frame (FPR).

\begin{equation}
\mbox{RBDR} = \frac{\mbox{num. of true positive regions (NTP)}}{\mbox{total num. of anomalous regions (TAR)}}.
\end{equation}
The RBDR is computed over all ground truth anomalous regions in all
frames of the test set.
\begin{equation}
\mbox{FPR} = \frac{\mbox{num. of false positive regions (NFP)}}{\mbox{total frames}}
\end{equation}
where FPR is the false-positive rate per frame.

The number of true positive regions (NTP) can be expressed as
\begin{equation}
\mbox{NTP} =\sum_{t=1}^T \sum_{i=1}^{N_t} [ \exists \; D^t \; \mbox{such that} \; \frac{G_i^t \cap D^t}{G_i^t \cup D^t} \geq \beta]
\end{equation}
where $D^t$ is a detected region in frame $t$, $G_i^t$ is the $i^{th}$ ground truth anomalous region in frame $t$, $N_t$ is the number of ground truth anomalous regions in frame $t$, and $\beta$ is a threshold which is set to 0.1 in \cite{ramachandra_street_2019}.

In other words, the number of true positive regions is the total number of ground truth regions in all testing frames that are detected.  A ground truth region in a frame is considered detected if the intersection over union (IOU) between the ground truth region and any detected region in the frame is greater than or equal to $\beta$.

The total number of ground truth anomalous regions can be expressed as
\begin{equation}
    \mbox{TAR} = \sum_{t=1}^{T} N_t
\end{equation}
where $N_t$ is the number of ground truth anomalous regions in frame $t$.

The number of false positive regions (NFP) can be expressed as
\begin{equation}
\mbox{NFP} =\sum_{t}^T \sum_{j=1}^{M_t} [ \forall \; G^t, \; \frac{G^t \cap D_j^t}{G^t \cup D_j^t} < \beta]
\end{equation}
where $G^t$ is a ground truth anomalous region in frame $t$, $D_j^t$ is the $j^{th}$ detected region in frame $t$, $M_t$ is the number of detected regions in frame $t$, and $\beta$ is a threshold set to 0.1. 

In other words, the number of false positive regions is the total number of detected regions in all testing frames that do not overlap enough with any ground truth anomalous region.

The other criteria introduced in \cite{ramachandra_street_2019}, the {\em track-based detection criterion}, measures the track-based detection rate (TBDR) versus the number of false positive regions per frame (FPR).  For this criterion, ground truth anomalous tracks are needed.  A ground truth anomalous track is a set of ground truth anomalous regions in a sequence of consecutive frames.
\begin{equation}
\mbox{TBDR} = \frac{\mbox{num. of true positive tracks (NTPT)}}{\mbox{total num. of anomalous tracks (NAT)}}.
\end{equation}
Without loss of generality, let us assume that in the notation $G_k^t$, $k$ further identifies an anomalous track. Then, an anomalous track can be defined as the set of ground truth anomalous regions it contains, spanning frames $t1$ to $t2$ as such:
$$L_k := \{G_k^{t1}, G_k^{t1 + 1}, ..., G_k^{t2 - 1}, G_k^{t2}\}$$
The number of true positive tracks can be expressed as 
\begin{equation}
\begin{split}
\mbox{NTPT} = \sum_{k = 1}^{N_k} & \left[ \left(\sum_{G_k^t \in  L_k} [ \exists \; D^t \; \mbox{such that} \; \frac{G_k^t \cap D^t}{G_k^t \cup D^t} \geq \beta]\right)\right. \\ 
& \left. \geq \alpha \cdot |L_k| \vphantom{\sum_{G_k^t \in  L_k}}\right]
\end{split}
\end{equation}
where $N_k$ is the total number of anomalous tracks (NAT), $|L_k|$ denotes the the size of $L_k$ and $\alpha$ is a threshold which is set to 0.1 in \cite{ramachandra_street_2019}.

In other words, a ground truth anomalous track is considered a true positive if at least a fraction $\alpha$ (set to 0.1) of the ground truth anomalous regions in the track are correctly detected.  The condition for detecting ground truth anomalous regions is the same as for the region-based criterion above.
\begin{equation}
\mbox{FPR} = \frac{\mbox{num. of false positive regions (NFP)}}{\mbox{total frames}}
\end{equation}
where FPR is the false-positive rate per frame.
A region predicted as anomalous in a frame is a false positive if the IOU between it and every ground truth region in that frame is less than $\beta$.  This is the same definition as for the region-based criterion.

Notice that since false positive regions are counted per frame, the maximum possible false positive rate for either criterion can exceed 1.0. The authors recommend summarizing the ROC curve by calculating AUC for false positive rates per frame from 0 to 1.0 for both criteria.

As a consequence of using these new criteria, bounding box annotations with unique anomaly IDs as well as track IDs are required, which the authors provide for the UCSD Ped1, UCSD Ped2, CUHK Avenue and Street Scene datasets.

Finally, one should also consider that measures such as AUC only provide a summary of a narrow view of performance, and have many drawbacks \cite{lobo_auc:_2008}. For these reasons, researchers are encouraged to provide qualitiative analysis and visualizations of detections. Of particular importance is the \textit{quality of false positives} predicted by different methods, which cannot possibly be captured without visual inspection. A method that produces false positives in test data corresponding to plausibly odd behaviors (that did not exist in the training data) should be favored to another that produces seemingly random false positives, when otherwise numerical measures such as AUCs are comparable between them.

\subsection{A Taxonomy of Video Anomaly Detection Approaches}
\label{subsect:taxonomy}
\begin{figure*}
    \centering
    \hspace*{5pt}
    \includegraphics[width=\linewidth]{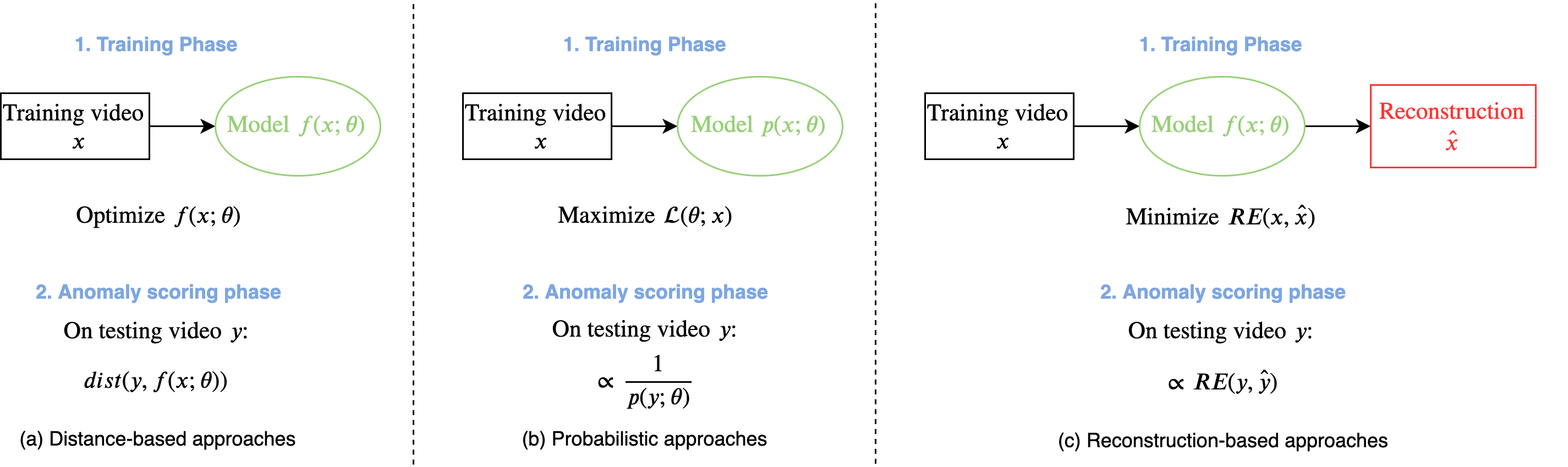}
    \vspace{-10pt}
    \caption{An overview of the 3 basic approaches past work has taken to video anomaly detection.}
    \vspace{-5pt}
    \label{fig:VAD-approaches}
\end{figure*}

At a high level, past video anomaly detection work can be categorized into \textbf{distance-based}, \textbf{probabilistic} and \textbf{reconstruction-based} approaches. See Figure \ref{fig:VAD-approaches} for intuition on how these approaches work and the subtle similarities and differences between them. Here we review popular works that evaluate performance on at least one of the aforementioned video anomaly detection benchmark datasets, but also give some treatment to seminal works in the area. These approaches are not mutually exclusive, as methods that seem to operate in a distance-based fashion at first sight could easily have probabilistic interpretations; the categorization is merely for convenience. Based off of the basic intuition behind video anomaly detection as explained in Figure \ref{fig:overview}, we further group methods by both the representation and modeling strategies they employ.

\subsubsection{Broad Themes in Representation}
\label{subsubsect:repr-themes}
Broadly, there are two classes of representations used by video anomaly detection approaches, hand-crafted features and deep features from a CNN. Hand-crafted features include spatio-temporal gradients (\cite{lu_abnormal_2013,ionescu_detecting_2019}),
dynamic textures (\cite{weixin_li_anomaly_2014,mahadevan_anomaly_2010}),
histogram of gradients (\cite{hasan_learning_2016,saligrama_video_2012,ma_anomaly_2015}), histogram of flows
(\cite{hasan_learning_2016,saligrama_video_2012,cong_abnormal_2013,cheng_video_2015}),
flow fields (\cite{adam_robust_2008,antic_video_2011,antic_spatio-temporal_2015,wu_chaotic_2010,mehran_abnormal_2009}), dense trajectories (\cite{ma_anomaly_2015})
and foreground masks (\cite{antic_video_2011,ramachandra_street_2019}). The deep features are further either extracted as-is from a pre-trained network (such as \cite{battiato_deep_2017,ionescu_detecting_2019,hinami_joint_2017,ravanbakhsh_plug-and-play_2018,luo_revisit_2017,sabokrou_deep-anomaly:_2018}) or are learned while optimizing for a particular task related to anomaly detection, such as with auto-encoders optimizing for low reconstruction error (such as \cite{hasan_learning_2016,ionescu_object-centric_2019,chong_abnormal_2017,sabokrou_deep-cascade:_2017,xu_learning_2015,sabokrou_real-time_2015,sabokrou_adversarially_2018}).

Another consideration in representation is the atomic unit of processing. Algorithms process atomic units ranging from image patches (such as \cite{antic_video_2011,cong_abnormal_2013,xu_learning_2015}) to video patches (such as \cite{weixin_li_anomaly_2014,lu_abnormal_2013,adam_robust_2008,ramachandra_street_2019,ionescu_object-centric_2019,mahadevan_anomaly_2010,ramachandra_learning_2019,ionescu_detecting_2019,sabokrou_deep-cascade:_2017,saligrama_video_2012,cheng_video_2015,mehran_abnormal_2009,luo_revisit_2017,sabokrou_real-time_2015,kim_observe_2009,benezeth_abnormal_2009,kratz_anomaly_2009}) to single full frames (such as \cite{battiato_deep_2017,hinami_joint_2017,ravanbakhsh_abnormal_2017}) and even video snippets (short sequences of full frames) (such as \cite{liu_future_2018,hasan_learning_2016,chong_abnormal_2017,ma_anomaly_2015,wu_chaotic_2010,ravanbakhsh_plug-and-play_2018,sabokrou_deep-anomaly:_2018}). When dealing with image or video patches, algorithms operate on units from single fixed-size patches (such as \cite{adam_robust_2008,ramachandra_street_2019,saligrama_video_2012}) to multi-scale fixed-size patches (such as \cite{antic_video_2011,antic_spatio-temporal_2015,xu_learning_2015}) to arbitrarily-sized region proposals (such as \cite{ramachandra2019anomaly}).

\subsubsection{Broad Themes in Modeling}
\label{subsubsect:modeling-themes}
Broad themes in modeling include the use of one-class SVMs (such as \cite{battiato_deep_2017,ionescu_detecting_2019,ma_anomaly_2015,xu_learning_2015}), nearest neighbor approaches (such as \cite{saligrama_video_2010,ramachandra_street_2019,ramachandra_learning_2019,cheng_video_2015,hinami_joint_2017}), Hidden Markov Models (such as \cite{weixin_li_anomaly_2014,kim_observe_2009,benezeth_abnormal_2009,kratz_anomaly_2009}) and more generally Probabilistic Graphical Models (such as \cite{weixin_li_anomaly_2014,antic_video_2011,antic_spatio-temporal_2015}). More recently, deep learning approaches have started using adversarial training strategies (such as \cite{liu_future_2018,sabokrou_adversarially_2018,ravanbakhsh_abnormal_2017}).

Some works focus solely on \textbf{frame-level (temporal) localization}, and in most cases, this means that this objective is built into the model, and as such, the models fail to perform adequate spatial localization (\cite{sultani_real-world_2018,hasan_learning_2016,chong_modeling_2015}). For instance, methods that use video snippets as their atomic unit of processing often also have a temporal detection focus.

Some works do not specifically account for the \textbf{location-dependent nature of anomalies}, such as \cite{liu_future_2018,hasan_learning_2016,ionescu_object-centric_2019,chong_abnormal_2017,cong_abnormal_2013,luo_revisit_2017,xu_learning_2015,ravanbakhsh_abnormal_2017}. For example, methods that use full frames or video snippets as their atomic unit of processing often overlook this characterization. That is, these methods would not be able to distinguish loitering outside an embassy building from loitering in a public park beside it; they operate under a looser definition of anomalies than that provided in Definition \ref{defn:video-anomalies}. 
Others account for the location-specific nature of anomalies in one of two-ways: (1) scoring voxels conditioned on their location in the camera frame (such as \cite{adam_robust_2008,ramachandra_street_2019,ramachandra_learning_2019,saligrama_video_2012}), (2) providing additional context in the form of information from neighboring voxels for scoring (such as \cite{weixin_li_anomaly_2014,antic_spatio-temporal_2015,kim_observe_2009}).

Another problematic practice that has emerged is that of \textbf{per-video normalization}, such as that in \cite{liu_future_2018,hasan_learning_2016,chong_abnormal_2017,luo_revisit_2017,ravanbakhsh_abnormal_2017,gong_memorizing_2019,park2020learning}. Here, for every testing sequence, abnormality scores are assigned per frame and subsequently min-max normalized using scores \textit{within the same video}. This practice has the inherent assumption that every test sequence has at least one normal and one anomalous frame. This is further problematic because scores assigned to frames across videos are not comparable anymore and this does not reflect the way real unseen data would have to be scored - the ``end of a test video sequence'' is unknown in practice.

\subsection{Other Less Common Settings and Related Modeling}
\label{subsect:less-common-settings}
\subsubsection{Object detection and tracking approach}
\label{subsubsect:detect-and-track}
This approach relies on being able to detect and track objects across time, creating complete object trajectories. Unfortunately, natural scenes rarely have the property that only certain object types that can be detected, such as humans (\cite{morais2019learning,rodrigues2020multi,markovitz2020graph}) or cars, are present. Moreover, natural scenes almost always present with occlusions. So while object detection and tracking methods would have vast utility in anomaly detection with the ability to detect trajectory anomalies, as of the current state of the art in computer vision, this is a clearly suboptimal approach for general video anomaly detection. 

For a specific example, consider the work of Morais et al. \cite{morais2019learning}. The authors take a human-detection-and-tracking reconstruction-based approach. They extract fixed-length tracklets for skeletons estimated by Alpha Pose \cite{fang2017rmpe} and decompose them into global and local components based on their properties. They design a global + local two branch architecture, each of which contains three GRUs \cite{cho2014learning} - an encoder, a reconstructing decoder and a predicting decoder. The global and local branches pass messages to each other about their states and the whole architecture is jointly optimized for reconstruction and future movement prediction through mean squared error terms on perceptual, global and local loss terms.
While effective for detecting anomalous human actions, their method is not applicable to the general video anomaly detection problem.  In their experiments, they exclude those sequences from ShanghaiTech and CUHK Avenue where the main subject is non-human or cannot be detected and tracked.

\subsubsection{Supervised anomaly detection}
\label{subsubsect:supervised}
Supervised approaches assume that anomalous data is available during training time. There are two main problems with this assumption: (1) all possible future anomalous activities cannot possibly be available and annotated in any natural scene, especially given that they occur so rarely and (2) even if all possible anomalous activities were available for supervision, the problem itself would reduce to binary video classification where the anomalous class is ``known''. This defeats the spirit of video anomaly detection where the ultimate goal in practice is to detect \textit{any} deviation from normality.

For a specific example, consider the work of \cite{zhou_spatialtemporal_2016}. The authors build a simple spatio-temporal CNN classifier to perform the normal/anomaly classification on fixed-size video patches. To achieve this, they use data from both the normal and anomalous classes for training.

\subsubsection{Video-level weak supervision}
\label{subsubsect:video-level-weak-sup}
This approach relies on having weak supervision in the training of a model in the form of video-level labels (as opposed to snippet/frame-level labels). As far as we know, this approach was mainly born out of the introduction of the UCF-Crime dataset which has video-level labels in both the training and test sets. While it has utility, this problem formulation seems to be an overly specific one. The immediate concern is as with the supervised setting: how can one expect to have videos of all possible anomalous activities at training time when they occur so rarely and are susceptible to concept drift?

The authors of \cite{sultani_real-world_2018} along with the presentation of the UCF-Crime dataset, discuss a Multiple Instance Learning (MIL) Framework for performing anomaly detection using weak supervision of this form. Bags contain fixed-length snippets of videos, where positive bags contain at least one anomalous snippet and negative ones none. They perform MIL ranking by enforcing a constraint that the maximum score over snippets in a positive bag must be greater than a negative bag and add additional sparsity and temporal smoothness constraints to provide better priors to the classification task. They present the first method that utilizes video-level labels for video anomaly detection, so they are able to compare only against methods that cannot utilize these labels. Zhu~\cite{zhu_motion-aware_2019} follows up on this work; they learn a motion-aware feature and demonstrate that it can provide large gains in the MIL framework.
Very recently, in \cite{zhong_graph_2019}, the authors convert this weakly-supervised formulation to a fully supervised one with noisy labels, where the primary task becomes to clean the noise and the secondary task of video anomaly detection is converted to binary video action recognition. They use a graph convolutional network \cite{kipf2016semi} to clean the noise in an alternating optimization mechanism.

\section{Distance-based Approaches}
\label{sect:dist-based}

Distance-based approaches involve using the training data to create a model of ``normality'' and measuring deviations from this model to determine anomaly scores. Usually, these models are themselves quite simple, but clever representation and formulation lead to good performance. From Figure \ref{fig:VAD-approaches} (a), distance-based approaches can be seen as a more general form of both probabilistic and reconstruction-based approaches.

Many different types of features have been used in distance-based approaches as well as many different ways of measuring distance to the normal features.  One common approach for many methods is to use one-class SVMs to compute a decision boundary around feature vectors from normal training video (such as \cite{ionescu_object-centric_2019,battiato_deep_2017,ionescu_detecting_2019,ma_anomaly_2015,xu_learning_2015,tran_anomaly_2017}).  A disadvantage of such approaches is that it is expensive to update the model given new normal training data.  The SVM learning algorithm must be rerun on all of the old plus new data.  An alternative approach is to use a mixture of Gaussians to model normal feature vectors and then Mahalanobis distance to measure distance to normality.  This idea has been used in the works \cite{sabokrou_deep-cascade:_2017,sabokrou_deep-anomaly:_2018,sabokrou_real-time_2015,sun_online_2017}.

In terms of the features used to represent video volumes, early approaches used a variety of hand-crafted features including foreground masks \cite{ramachandra_street_2019,saligrama_video_2012}, histograms of flow \cite{dalal2006human}, motion magnitude \cite{saligrama_video_2012}, histogram of gradients \cite{dalal2005histograms}, motion boundary histograms \cite{laptev2008learning}, dense trajectories \cite{wang2013dense}, and space time interest point (STIP) features \cite{dollar2005behavior}.  More recent approaches in \cite{ramachandra_street_2019,ionescu_object-centric_2019,battiato_deep_2017,ionescu_detecting_2019,ravanbakhsh_plug-and-play_2018,sabokrou_deep-anomaly:_2018,xu_learning_2015,tran_anomaly_2017} have focused on features learned by deep networks which generally have higher accuracy (see Table \ref{tab:auc-table}).  These deep-network-based methods encompass a variety of ways of learning deep features and a variety of ways of using the deep features in different models of normality. 

The remainder of this section briefly summarizes a number of distance-based approaches.

In \cite{saligrama_video_2012}, the main premise is that \textbf{anomalies have local spatio-temporal ``signatures''}, causing them to have low likelihood under a joint probability distribution of local normal data. They extract overlapping fixed-size video volumes and represent them with low-level motion descriptors.  Aggregated K-nearest neighbors (K-NN) distances between normal and test video volumes are used to compute anomaly scores.


In \cite{ma_anomaly_2015}, the authors extract a set of social force \cite{helbing1995social}, HOG (histogram of gradients) \cite{dalal2005histograms}, HOF (histogram of optical flow) \cite{dalal2006human}, MBH (motion boundary histogram) features \cite{laptev2008learning} and dense trajectories \cite{wang2013dense} from video snippets. They use \textbf{Vector Quantization (VQ) coding} to represent the features and a \textbf{one-class SVM} \cite{scholkopf2000support}, with either linear, RBF or polynomial kernels to perform anomaly detection.

In \cite{xu_learning_2015}, the authors propose \textbf{one of the first approaches that used \textit{learned} representations with deep networks for video anomaly detection}. They use two streams (RGB and optical flow) of stacked denoising auto-encoders (DAE) on multi-scale fixed-size overlapping video volumes to learn low-dimensional representations. They then use the latent codes from the DAEs in a one-class SVM \cite{scholkopf2000support} with an RBF kernel to perform one-class classification for anomaly detection. They further present two ways to perform fusion between the modalities, at the representation stage and at the scoring stage.

In \cite{sabokrou_real-time_2015}, the authors focus on a fast method for video anomaly detection.  They use \textbf{Gaussians to model the distributions of features} from a simple 2-layer auto-encoder as well as the distribution of distances of each video volume to its spatio-temporal neighbors using a Structural Similarity Index Measure (SSIM).  They detect anomalies by computing the Mahalanobis distances to the training Gaussians.


Sabokrou et al. \cite{sabokrou_deep-cascade:_2017} build on their earlier work in \cite{sabokrou_real-time_2015}.  They use many internal layers of a 3D auto-encoder as well as a deep CNN to provide features which are modeled by Gaussian distributions in a cascade structure.  

This work was followed by \cite{sabokrou_deep-anomaly:_2018} which simplifies the cascade architecture into a two-layer cascade of Gaussian models and uses features from a pre-trained fully convolutional network. The resulting algorithm can process hundreds of video frames per second on a high-end GPU.


Smeureanu et al. \cite{battiato_deep_2017} present \textbf{one of the first approaches that makes use of features from a pre-trained CNN} for video anomaly detection. This is one of the only approaches to use single frames as their atomic processing unit. They train a one-class SVM \cite{scholkopf2000support} with a linear kernel on deep features extracted from a VGG-f network on each mean-subtracted frame \cite{chatfield2014return}. They smooth their score maps with a spatio-temporal filter and perform localization by dividing video into fixed-size video volumes and simply aggregating anomaly scores over the patch regions.

In \cite{tran_anomaly_2017}, the authors present a way to use \textbf{convolutional winner-take-all auto-encoders} \cite{makhzani2015winner} to learn motion-feature representations from optical flow fields of fixed-size video volumes. They then use the learnt motion-feature representations to build location-dependent one-class SVMs \cite{scholkopf2000support} to perform anomaly scoring.

In \cite{battiato_convex_2017}, the authors present a \textbf{unique geometric approach} to anomaly detection. They use dense trajectories from training frames to create an ensemble of extended convex hulls \cite{casale2014approximate}, identifying anomalies at test time using a \textbf{polytope inclusion test}, presumably scoring individual trajectories using their distance-to-convex-hull. They also cluster potentially anomalous trajectories to detect anomalous regions and filter out small false positive detections.

In \cite{sun_online_2017}, the authors build a model of normality using the \textbf{Growing Neural Gas} \cite{fritzke1995growing} algorithm on STIP features \cite{dollar2005behavior} extracted from video snippets/volumes. They contend that past methods have not sufficiently dealt with ``changing scenes'' and propose augmenting the GNG model with online updates in the form of neuron insertion, deletion, learning rate adaptation and stopping criteria. Detection is performed by simply determining whether new patterns are significantly different from nearest-neighbor in the GNG model by studying the distribution of distances.


In \cite{ravanbakhsh_plug-and-play_2018}, the authors present another way to use image features from a pre-trained convolutional network, AlexNet \cite{krizhevsky2012imagenet}. They also propose a two-stream model, operating on both appearance features and optical flow fields. Using the CNN-extracted features, they apply \textbf{Iterative Quantization Hashing} \cite{gong2012iterative} via a pre-trained binary fully convolutional network to generate binary maps for each frame. They then develop a \textbf{Temporal CNN Pattern (TCP) measure}, a statistical measure of the amount of change of the appearance features over time. Fusion of the two streams produces their final anomaly score maps.

In \cite{sabokrou_adversarially_2018}, the authors present \textbf{one of the first approaches to use adversarial training} for video anomaly detection. They use a discriminator network ($\mathcal{D}$) tasked with distinguishing original image patches from reconstructions of noisy patches obtained from a denoising auto-encoder network ($\mathcal{R}$) which plays the role of generator. Since $\mathcal{R}$ is trained only on image patches from training data, it decimates outliers and thus enables $\mathcal{D}$ to tell an anomalous image patch from its reconstruction easily.

In \cite{ionescu_detecting_2019}, the authors propose a two-stage anomaly detection algorithm. They extract fixed-size video volumes from training video, augment them with location, appearance (extracting feature maps from a pre-trained CNN) and motion information (in the form of 3D gradients). For first stage detection, they perform $K$-means clustering and eliminate small clusters corresponding to noise/outliers to create a robust representation. Second stage detection involves building $K$ one-class SVMs (one for each cluster) to create a \textbf{``narrowed normality clusters''} model, and at test time treating the maximum score for a test patch under these $K$ one-class SVMs as the abnormality score.

In \cite{ionescu_object-centric_2019}, the authors \textbf{convert the anomaly detection problem to $k$ multi-class 1-versus-rest classification problems}, building on their previous work \cite{ionescu_detecting_2019}. They use feature pyramid networks \cite{lin2017feature} to extract crops, train convolutional auto-encoders on appearance and gradient features of these crops to learn latent representations and then perform $k$-means clustering followed by training of $k$ one-class SVMs to make binary one-versus-rest classifications. At test time they simply use the inverse of the maximum of $k$ classification scores as an anomaly score. They do not report spatial localization performance.

In \cite{ramachandra_street_2019}, the authors present two baseline algorithms for future comparison on their recently released dataset, Street Scene. They use a simple \textbf{nearest neighbor location-dependent anomaly detection} scheme using hand-crafted representations of video volumes (flow fields or blurred foreground masks) along with hand-crafted distance measurement (a normalized L1 or L2 voxel-wise distance respectively). They vastly reduce the number of distance computations by building a concise representative \textbf{exemplar model from training data}. Interestingly, they show that these simple methods are able to outperform some of the previous state of the art methods on other datasets, possibly indicating that algorithms have developed biases specific to certain datasets.

In \cite{ramachandra_learning_2019}, the authors build on the simple nearest neighbor scheme by replacing the hand-crafted representation and distance function with \textit{learned} ones by training a \textbf{Siamese neural network} \cite{bromley_signature_1994}. The Siamese network is trained to classify video patch pairs as similar or different and is used to find testing video volumes that are different from all training video volumes and are therefore anomalous.  An exemplar model (consisting of all unique normal video volumes) is learned from training data of the target dataset.  Finally, nearest neighbor scoring between test video volumes and exemplars using the trained Siamese network is used to assign anomaly scores to each testing video patch.

\section{Probabilistic Approaches}
\label{sect:stats}

Probabilistic approaches compute distance under a model in some probability space. These methods usually aim to admit modeling into a probabilistic framework such as with probabilistic graphical models (PGMs) or high-dimensional mixtures of probability distributions. See Figure \ref{fig:VAD-approaches} (b) for this intuition.
Most of the probabilistic approaches came before the wave of deep learning methods and instead rely on features such as spatio-temporal gradients \cite{kratz_anomaly_2009}, optical flow fields \cite{adam_robust_2008,antic_video_2011,antic_spatio-temporal_2015,wu_chaotic_2010,mehran_abnormal_2009,kim_observe_2009}, and STIP features \cite{cheng_video_2015} coupled with traditional models such as Markov Random Fields \cite{kim_observe_2009,benezeth_abnormal_2009,kratz_anomaly_2009} and mixtures of Gaussians \cite{wu_chaotic_2010,kratz_anomaly_2009}.  A couple of more recent approaches do make use of deep networks \cite{hinami_joint_2017,feng_learning_2017} and also show improved accuracy. These approaches enjoy a favorable property in being highly principled and having the ability to model the continuous nature of anomalousness well. Unfortunately, they are often very slow at test time. We summarize the various probabilistic approaches below.

In \cite{adam_robust_2008}, the authors use \textbf{fixed-location monitors} on the camera frame which have a fixed-size storage buffer in which they store optical flow fields. They declare anomalies as those test optical flow observations with low likelihood given the corresponding monitor's buffer, which they model either as a histogram of observations or using kernel density estimation.

In \cite{mehran_abnormal_2009}, the authors utilize the \textbf{social force} model \cite{helbing1995social}. Optical flow is used to estimate social force interactions which are roughly the difference between a pixel's optical flow and the average optical flow in a neighborhood around the pixel.  The idea being that the reason a pixel differs from its neighbors is due to interactions among particles.  A bag-of-words model is used to model social force interactions and anomalies are detected as low-likelihood frames under the model.


In \cite{benezeth_abnormal_2009}, the authors compute \textbf{binary motion labels for each pixel} by simple background subtraction. They use spatio-temporal neighborhoods around each pixel to compute co-occurrence statistics on the motion label representation of normal data and use the \textbf{co-occurrence matrix} as the potential function in a Markov Random Field to perform anomaly detection via likelihood ratio testing.

In \cite{kratz_anomaly_2009}, the authors represent video with spatio-temporal gradients. They use multivariate Gaussians to model their distribution for video patches and a mixture of Gaussians to represent the distribution of video patches for a given location in the camera frame. Finally, they use a \textbf{coupled Hidden Markov Model} to incorporate the effect of spatial and temporal correlations between the video patches.

Kim et al. \cite{kim_observe_2009} present a way to use a spatio-temporal Markov Random Field to model relationships between neighboring training video patches extracted from a grid on video. They \textbf{represent each video patch as a node in the graph} by building a Mixture of Probabilistic Principal Components Analyzers (MoPPCA) \cite{tipping1999mixtures} on optical flow observations. They detect anomalies by computing a maximum a posteriori estimate of normality at test time. They also show how their model can be incrementally updated to account for environmental changes and concept drift.

In \cite{wu_chaotic_2010}, the authors also advect particles on a grid of optical flow observations on video similar to \cite{mehran_abnormal_2009}, but they focus on trajectories of these particles. They cluster these trajectories and model chaotic dynamics of them using two \textbf{chaotic invariants}. Anomaly detection is performed by simply estimating parameters of a Gaussian mixture model on this chaotic feature set from normal data and evaluating the likelihood of test data.

In \cite{mahadevan_anomaly_2010}, the authors propose learning a \textbf{Mixture of Dynamic Textures} (MDT) \cite{doretto2003dynamic,chan2008modeling} from training video patches, with the mixtures shared across larger ``cell'' regions. They detect anomalies as those regions with high center-surround saliency as given by a discriminant saliency criterion \cite{gao2009decision}. In \cite{weixin_li_anomaly_2014}, they build off the MDT representation to operate at multiple scales. They integrate spatial and temporal anomaly scores from multiple scales using a conditional random field \cite{lafferty2001conditional} framework.

The authors in \cite{antic_video_2011} use a rather unique premise - that \textbf{anomaly detection must be done indirectly by trying to ``explain away'' the normality} in the test data using information learned from the training data. They seek a \textbf{video parsing} approach that simultaneously discovers foreground object hypotheses that jointly explain the foreground in a frame and those that have matching normal exemplar hypotheses. Those object hypotheses at test time which are necessary to explain the foreground but do not match any exemplar hypotheses from normal training data are anomalous. In \cite{antic_spatio-temporal_2015}, they further build on this idea by considering object hypotheses in the form of \textbf{flexible video pipes instead of just image patches}.

In \cite{cheng_video_2015}, the authors propose a hierarchical local plus global method to detect anomalies. They model video with Spatio-Temporal Interest Point (STIP) features \cite{dollar2005behavior} and form a codebook with K-means clustering, detecting local anomalies as those with high distance to the $k^{th}$ nearest neighbor. For global anomalies, they consider ensembles of STIP features to construct a high-level codebook of interaction templates and build \textbf{Gaussian Process Regression} (GPR) models \cite{rasmussen2003gaussian} with an RBF kernel for each model. They then designate low-likelihood test ensembles under the $k^{th}$ nearest neighboring GPR ensemble model as anomalous.

In \cite{hinami_joint_2017}, the authors propose a \textbf{unique method to recount anomalous events as they are detected}. They first train a Fast-RCNN \cite{girshick2015fast} model to predict object, action and attribute classes from large-scale COCO \cite{lin2014microsoft} and Visual Genome \cite{krishna2017visual} image datasets. Then for each frame, they extract features of each region of interest (RoI) from the second-to-last fully connected layer and perform anomaly detection with either nearest neighbor distance to training sample, a one-class SVM with RBF kernel or likelihood under a kernel density estimate with RBF kernel. Recounting is performed by simply looking at maximal predictions of object, action and attribute classes. 

In \cite{feng_learning_2017}, the authors use PCANet \cite{chan2015pcanet} to extract deep representations, learned from 3D gradients of normal image patches. They then use \textbf{Deep GMMs} \cite{van2014factoring} \textbf{to model a generative process of normal patterns}, maximizing a lower-bound on log-likelihood. The deep GMM model simply yields likelihood scores for testing patterns which are used as anomaly scores.

\section{Reconstruction-based Approaches}
\label{sect:recon-based}

Reconstruction approaches aim to represent the input (images or video snippets) using a high-level or compact representation learned from normal video and then reconstruct the input using only this representation.  They are based on the premise that out-of-distribution inputs such as anomalies are inherently harder to reconstruct using a representation learned from normal video when compared to in-distribution normal data, thus justifying the use of reconstruction error as a proxy for anomaly score. See Figure \ref{fig:VAD-approaches} (c) for an illustration of this intuition.  Almost all of the reconstruction-based approaches use modern deep learning methods, and in particular, most are based on either convolutional auto-encoders \cite{hasan_learning_2016,chong_abnormal_2017,vu_robust_2019,gong_memorizing_2019} or generative adversarial networks (GANs) \cite{vu_robust_2019,ravanbakhsh_abnormal_2017,nguyen_anomaly_2019,tang_integrating_2020}.  Generally, reconstruction-based approaches have the disadvantage that the models they use (e.g. auto-encoder or GAN) need to be retrained to accomodate new normal training video.  Many of these approaches do not evaluate spatial localization of anomalies despite the fact that reconstruction error is generally pixelwise. Presumably, this is because their spatial localization accuracy is low. Another disadvantage of auto-encoder-type reconstruction methods is that reconstruction errors for frames are proportional to the number of foreground objects in the frame and this is the reason most of these methods have to employ a post-processing step of per-video normalization, as previously discussed. We summarize the various reconstruction-based approaches in the remainder of this section.

In \cite{hasan_learning_2016}, the authors train a \textbf{convolutional auto-encoder} to reconstruct training video snippets with a pixel-wise L2 loss. Reconstruction error on testing video snippets, normalized per-video sequence, serves as their abnormality scores, They do not perform spatial localization, claiming a focus on temporal localization. Interestingly, they also train a generalized auto-encoder on training data from several datasets and show that it performs about as well as one trained on a single dataset. Rather than demonstrating robustness of features, we believe this actually indicates a common bias towards anomalous activities being caused by objects with faster motion in the group of datasets they perform their experiments on.

Chong et al. \cite{chong_abnormal_2017} build on the convolutional auto-encoder architecture of \cite{hasan_learning_2016} by preserving temporal ordering of frames through the convolutions and modeling the temporal information at the bottleneck layer with specialized \textbf{convolutional LSTM} \cite{xingjian_convolutional_2015} layers.

In \cite{ravanbakhsh_abnormal_2017}, the authors attempt the \textbf{first use of Generative Adversarial Networks} (GANs) \cite{goodfellow2014generative} for video anomaly detection. They train two conditional GANs, that take as input $(x, z)$ pairs of frames and noise vectors and generate corresponding frames $y$ of a different modality (they use raw frames to optical flows and vice versa in the two GANs). The discriminators are asked to classify pairs of $(x, y)$ representations of frames as real or fake. Assuming that anomalies are not reconstructed well, they fuse reconstruction errors from both modalities, and use per-video normalization to perform anomaly scoring for detection and pixel-wise localization.

In \cite{vu_energy-based_2017}, the authors perform feature learning and reconstruction on fixed-size raw video patches using \textbf{Restricted Boltzmann Machines} (RBMs) \cite{freund1992unsupervised} using the Contrastive Divergence \cite{hinton2002training} training algorithm. They combine reconstruction errors at test time from different pyramid levels and overlapping patches to come up with an anomaly score.

In \cite{liu_future_2018}, the authors contend that \textbf{predicting a video snippet's future frame must be harder for anomalous activities compared to normal ones}, and thus design a future frame prediction framework. They train a U-net-style network \cite{ronneberger2015u} that takes training video snippets of length $t$ as input and predicts a future frame for time $t + 1$. Further, they use FlowNet \cite{dosovitskiy2015flownet} to estimate pairs of optical flow maps between the frame at $t$ and real or reconstructed frames at $t + 1$. L1 losses between flow maps, intensity and directional gradients of reconstructions along with an adversarial loss to differentiate the real and reconstructed frames at $t + 1$, followed by per-video normalization of errors, forms their anomaly score. They also do not report spatial localization performance.

In \cite{nguyen_anomaly_2019}, the authors address the problem by \textbf{learning a correspondence between common object appearances and their associated motions} in a two-stream model. Using a single frame as input, they use a single encoder coupled with both a U-net decoder that predicts motion as well as a devoncolutional decoder that reconstructs the input frame, governed by $l_p$ reconstruction error loss terms. They consider this entire network a generator in a conditional GAN, where the discriminator is another small network that distinguishes between pairs of input frames and corresponding real/estimated flow fields which is governed by a binary classification loss.  For testing frames, they calculate $l_p$ scores at a patch-level and use per-video normalization of scores for their final frame-level anomaly scores. They also do not report spatial localization performance.

In \cite{vu_robust_2019}, the authors observe that past reconstruction-based methods have largely operated on low-level features. They seek to address this by performing anomaly detection only with abstract features. First, they train \textbf{Denoising Auto-encoders} (DAEs) on raw video snippets and corresponding flow field representations. They then extract representations at multiple layers and train \textbf{conditional GANs} for each similar to \cite{ravanbakhsh_abnormal_2017}. Lastly, they combine reconstruction error maps from the multiple levels to arrive at a consensus score map for each frame.

In \cite{tang_integrating_2020}, the authors contend that \textbf{prediction and reconstruction can be combined to exploit advantages and balance disadvantages of both}. They seek to do this by creating a generator that operates on video snippets comprised of two consecutive U-net \cite{ronneberger2015u} architectures, where the first predicts an intermediate ``frame'' that is then used by the second to predict the immediate future frame, trained end-to-end by minimizing reconstruction error on intensity and gradient modalities. They also employ an adversarial loss on either ground truth future and predicted future frame pairs or at a finer level similar to PatchGAN \cite{isola2017image}.

In \cite{park2020learning}, the authors contend that CNN-based reconstruction approaches suffer from reconstructing anomalous events well because of CNNs' high representational capacity. They propose augmenting a U-net style encoder-decoder future frame prediction/reconstruction network with a \textbf{learned memory module} that stores important normal patterns and computing anomaly scores using a combination of PSNR between a frame and its reconstruction as well as distance between an encoding and nearest memory element. They also perform per-video normalization of scores.

\subsection{Sparse Reconstruction Approaches}
\label{subsect:sparse-recon}
A subset of reconstruction approaches, sparse reconstruction approaches, impose an additional constraint on the reconstruction in that it must be performed using a sparse feature set only. Almost all sparse reconstruction approaches optimize a sparse combination learning formulation of some kind \cite{lu_abnormal_2013,lu2019fast,cong_abnormal_2013}. These approaches usually enjoy some favorable properties in being fast (since sparsity is a goal) and having models of normality that are easy to update in an online fashion. A disadvantage of these approaches is that they often rely too heavily on \textit{memorizing} salient normal features, placing a large burden on the normal training set being exhaustive. They also do not tend to model the intuition behind the continuous nature of anomalousness very well due to this, that is, anomalies that received smaller scores often do not necessarily correspond to less anomalous activities as per human intuition.

In \cite{cong_abnormal_2013}, the authors estimate optical flow fields in video and extract a multi-scale histogram of flow features. They then learn a \textbf{dictionary} of these features from training video and use a \textbf{sparse reconstruction cost} based on L1 minimization as their anomaly score on volumes from test video. Favorable properties include online update of the dictionary and the ability to define bases over different representations such as image patches, video volumes or video snippets to perform anomaly detection at various levels.

In \cite{lu_abnormal_2013}, the authors operate on 3D gradient features of fixed-size video patches extracted from video at multiple scales. They propose \textbf{sparse combination learning} from training video, where the goal is to learn a dictionary of atomic units from training video patches and sets of sparse combinations of these to reconstruct video patches. During test time, the sparse combination with the least reconstruction error is used to score test video patches. In \cite{lu2019fast}, the authors extend this work into a \textbf{birth-and-death combination online solver} to handle both dynamic and large-scale data. They also improve detection speed from 150 FPS to an impressive 1000 FPS.

In \cite{luo_revisit_2017}, the authors propose a temporally-coherent sparse coding (TSC) approach \textbf{with the constraint that temporally close frames be encoded with similar sparse coefficients}. They use a special type of stacked recurrent neural network (sRNN) to enforce this and by optimizing all parameters of this network simultaneously, avoid the non-trivial hyperparameter search involved in TSC. Interestingly, the representations they operate on are multi-scale pooled features extracted from a pre-trained network on UCF-101 for each full frame.  Because they have a focus on temporal anomaly detection, their method does not perform localization.

In \cite{gong_memorizing_2019}, the authors propose \textbf{augmenting a 3D convolutional auto-encoder with a memory module}. They argue that this would help overcome some other auto-encoder approaches generalizing ``too well'' on test data leading to missed detections. At the bottleneck layer, they implement a memory module to use a fixed-size memory with attention-based addressing and hard shrinkage to encourage sparse reconstructions of input video snippets. They also perform per-video normalization and do not report spatial localization performance.

\section{A comparative study of methods}
\label{sect:comp-study}
Table \ref{tab:thematic-grouping} lists papers discussed in the previous sections grouped by the type of approach taken, ordered chronologically per approach. The table also summarizes the common types of representation used and the common characteristics of the model of normal activity used. The following list explains the abbreviations used in the table.

\begin{table*}[ht]
\centering
\caption{Thematic Grouping by Representation and Modeling Strategies Taken.}
\label{tab:thematic-grouping}
\resizebox{\textwidth}{!}{%
\begin{tabular}{l|l|ccc||cccc}
\hline
{\bf Method} & {\bf Approach} & \multicolumn{3}{c||}{\bf Representation theme} & \multicolumn{4}{c}{\bf Modeling theme}\\
                        &                           & {\bf Proc. unit}    & {\bf Input feats.}                  & {\bf Pre-trained net}       & {\bf Model component}           & {\bf Per-video norm.}       & {\bf Location-dependent}    & {\bf Sp-local.}  \\ \hline
                                           \cite{saligrama_video_2012} & Dist.                     & Fixed-size VP & HOF, fg\%, motion magnitude   &                       & NN              &                       & \cmark &                       \\ \hline
                                           \cite{ma_anomaly_2015} & Dist.                     & VS            & HOF, SF, dense trajectories   &                       & OC-SVM          &                       &                       &                       \\ \hline
                                           \cite{xu_learning_2015} & Dist.                     & Fixed-size IP & Raw, flow, deep               &                       & OC-SVM, AE      &                       &                       & \cmark \\ \hline
                                           \cite{sabokrou_real-time_2015} & Dist.                     & Fixed-size VP & Raw, SSIM, deep               &                       & AE              &                       &                       & \cmark \\ \hline
                                           \cite{sabokrou_deep-cascade:_2017} & Dist.                     & Fixed-size VP & Raw, deep                     &                       & AE              &                       & \cmark & \cmark \\ \hline
                                           \cite{battiato_deep_2017} & Dist.                     & FF            & Raw, deep                     & \cmark & OC-SVM          &                       &                       & \cmark \\ \hline
                                           \cite{tran_anomaly_2017} & Dist.                     & Fixed-size VP & Flow, deep                    &                       & OC-SVM, AE      &                       & \cmark & \cmark \\ \hline
                                           \cite{battiato_convex_2017} & Dist.                     & FF            & dense trajectories            &                       &                 &                       &                       & \cmark \\ \hline
                                           \cite{sun_online_2017} & Dist.                     & VS, VP        & STIP                          &                       & NN              &                       &                       & \cmark \\ \hline
                                           \cite{sabokrou_deep-anomaly:_2018} & Dist.                     & VS            & Raw, deep                     & \cmark & AE              &                       &                       & \cmark \\ \hline
                                           \cite{ravanbakhsh_plug-and-play_2018} & Dist.                     & FF, VS        & Flow, deep                    & \cmark &                 &                       & \cmark & \cmark \\ \hline
                                           \cite{sabokrou_adversarially_2018} & Dist.                     & Fixed-size IP & Raw, deep                     &                       & Adversarial, AE &                       &                       &                       \\ \hline
                                           \cite{ionescu_detecting_2019} & Dist.                     & Fixed-size VP & 3D grad., deep                & \cmark & OC-SVM          &                       & \cmark & \cmark \\ \hline
                                           \cite{ionescu_object-centric_2019} & Dist.                     & Fixed-size VP & 2D grad., deep                &                       & OC-SVM, AE      &                       &                       &                       \\ \hline
                                           \cite{ramachandra_street_2019} & Dist.                     & Fixed-size VP & Flow, fg-mask                 &                       & NN              &                       & \cmark & \cmark \\ \hline
                                           \cite{ramachandra_learning_2019} & Dist.                     & Fixed-size VP & Flow, deep                    &                       & NN              &                       & \cmark & \cmark \\ \hline \hline
                                           \cite{adam_robust_2008} & Prob.                     & Fixed-size VP & Flow                          &                       &                 &                       & \cmark & \cmark \\ \hline
                                           \cite{mehran_abnormal_2009} & Prob.                     & Fixed-size VP & Flow, social force            &                       &                 &                       &                       & \cmark \\ \hline
                                           \cite{benezeth_abnormal_2009} & Prob.                     & Fixed-size VP & Fg-mask, co-occurrence matrix &                       & HMM             &                       & \cmark & \cmark \\ \hline
                                           \cite{kratz_anomaly_2009} & Prob.                     & Fixed-size VP & 3D grad.                      &                       & HMM             &                       & \cmark & \cmark \\ \hline
                                           \cite{kim_observe_2009} & Prob.                     & Fixed-size VP & Flow                          &                       & HMM             &                       & \cmark & \cmark \\ \hline
                                           \cite{wu_chaotic_2010} & Prob.                     & VS            & Flow                          &                       &                 &                       & \cmark & \cmark \\ \hline
                                           \cite{mahadevan_anomaly_2010} & Prob.                     & Fixed-size VP & DT                            &                       &                 &                       & \cmark & \cmark \\ \hline
                                           \cite{antic_video_2011} & Prob.                     & Fixed-size IP & Fg-mask, flow                 &                       & OC-SVM          &                       & \cmark & \cmark \\ \hline
                                           \cite{weixin_li_anomaly_2014} & Prob.                     & Fixed-size VP & DT                            &                       & HMM             &                       & \cmark & \cmark \\ \hline
                                           \cite{antic_spatio-temporal_2015} & Prob.                     & Fixed-size VT & Fg-mask, flow                 &                       & OC-SVM          &                       & \cmark & \cmark \\ \hline
                                           \cite{cheng_video_2015} & Prob.                     & Fixed-size VP & STIP, 3DSIFT, HOG, HOF        &                       & NN              &                       & \cmark & \cmark \\ \hline
                                           \cite{zhang_combining_2016} & Prob.                     & Fixed-size VP & 3D grad., HOF                 &                       & OC-SVM          &                       & \cmark & \cmark \\ \hline
                                           \cite{hinami_joint_2017} & Prob.                     & FF            & Raw, deep                     & \cmark & NN, OC-SVM      &                       & \cmark & \cmark \\ \hline
                                           \cite{feng_learning_2017} & Prob.                     & Fixed-size IP & 3D grad., deep                &                       &                 &                       &                       & \cmark \\ \hline \hline
                                          \cite{hasan_learning_2016}  & Recon.                    & VS            & Raw, deep                     &                       & AE              & \cmark &                       &                       \\ \hline
                                           \cite{chong_modeling_2015} & Recon.                    & VS            & Raw, deep                     &                       & AE              & \cmark &                       &                       \\ \hline
                                           \cite{ravanbakhsh_abnormal_2017} & Recon.                    & FF            & Raw, flow, deep               &                       & Adversarial     & \cmark &                       & \cmark \\ \hline
                                           \cite{vu_energy-based_2017} & Recon.                    & Fixed-size VP & Raw, deep                     &                       &                 &                       & \cmark & \cmark \\ \hline
                                           \cite{liu_future_2018} & Recon.                    & VS            & Raw, flow, deep, 2D grad.     &                       & Adversarial     & \cmark &                       &                       \\ \hline
                                           \cite{nguyen_anomaly_2019} & Recon.                    & FF            & Raw, flow, deep               &                       & Adversarial, AE & \cmark &                       & \cmark \\ \hline
                                           \cite{vu_robust_2019} & Recon.                    & VS            & Raw, flow, deep               &                       & Adversarial, AE &                       &                       & \cmark \\ \hline
                                           \cite{tang_integrating_2020} & Recon.                    & VS            & Raw, flow, deep               &                       & Adversarial     &                       &                       & \cmark \\ \hline
                                           \cite{park2020learning} & Recon.                    & VS            & Raw, deep               &                       & AE     &  \cmark                     &                       &  \\ \hline
                                           \cite{cong_abnormal_2013} & S-Recon.                  & IP, VP, VS    & HOF. flow                     &                       &                 &                       &                       & \cmark \\ \hline
                                          \cite{lu_abnormal_2013}  & S-Recon.                  & Fixed-size VP & 3D grad.                      &                       &                 &                       & \cmark & \cmark \\ \hline
                                           \cite{luo_revisit_2017} & S-Recon.                  & Fixed-size VP & Deep                          & \cmark &                 & \cmark &                       &                       \\ \hline
                                           \cite{gong_memorizing_2019} & S-Recon.                  & VS            & Raw, deep                     & \cmark & AE              & \cmark &                       & \cmark \\ \hline
\end{tabular}%
}
\end{table*}

\vspace*{0.1in}
\noindent \textbf{Thematic grouping Table \ref{tab:thematic-grouping} notation guide:}
\begin{itemize}
\item Proc. unit: Atomic unit of processing.
\item VS: Video snippets.
\item FF: Full frames.
\item VP: Video patch.
\item IP: Image patch.
\item VT: (Flexible) video tube.
\item Input feats: Input feature representation.
\item grad.: gradients, 2D or 3D.
\item flow: Sparse or dense optical flow representation of the processing unit, without binning into histograms.
\item deep: deep features in some form, such as extracted from a pre-trained CNN or learned end-to-end.
\item HOG: Histogram of Oriented Gradients \cite{dalal2005histograms}.
\item HOF: Histogram of Optical Flow \cite{dalal2006human}.
\item MBH: Motion Boundary Histogram \cite{laptev2008learning}.
\item Dense trajectories: \cite{wang2013dense}.
\item Social Force:  \cite{helbing1995social}.
\item DT: Mixtures of Dynamic Textures \cite{doretto2003dynamic,chan2008modeling}.
\item STIP: Spatio-temporal Interest Point features \cite{dollar2005behavior}.
\item 3DSIFT: 3-dimsional Scale Invariant Feature Transform features \cite{scovanner20073}.
\item OC-SVM: Use of one-class SVM \cite{scholkopf2000support}.
\item NN: Use of nearest neighbor logic.
\item HMM: Use of a vanilla Hidden Markov Model or its more specialized variants such as Markov Random Fields or Conditional Random Fields.
\item Adversarial: Use of an adversarial training procedure in some form.
\item AE: Use of a vanilla auto-encoder or its more specialized variants such as variational, denoising, contractive, or sparse auto-encoders.
\item Per-video norm.: The method performs normalization of anomaly scores per test sequence, encoding an assumption that every test sequence contains at least one normal and one anomalous frame.
\item Location-dependent: Operates in a location-dependent fashion, local spatial context is considered when detecting anomalies. See Section \ref{sect:introduction}.
\item Sp-local.: The method is apparently able to perform spatial localization.
\end{itemize}

To compare various methods in terms of accuracy, we have compiled Tables \ref{tab:auc-table}, \ref{tab:track-region-aucs} and \ref{tab:SS-aucs} showing the accuracy of many algorithms on the various datasets and evaluation criteria discussed earlier. Although some datasets have evolved both in terms of size and annotations over a period of time, these changes have been modest and we believe that since these datasets are released with pre-defined training and testing splits, the numbers in these tables are a reliable measure of performances in the ways the evaluation criteria aim to capture them. Table \ref{tab:auc-table} compares different methods by their frame and pixel-level criteria scores on UCSD Ped1, UCSD Ped2 and CUHK Avenue datasets. Table \ref{tab:track-region-aucs} compares methods via track and region-based criteria for these datasets, highlighting that we may not be as close to saturating performance on these datasets as the traditional criteria might indicate. Table \ref{tab:SS-aucs} brings into perspective how some of the methods that perform very well on the traditional criteria on the small datasets display very poor performance on the large and complex Street Scene dataset, regardless of the evaluation criteria used.

Something that Table \ref{tab:auc-table} makes clear is that there currently is not a single best method.  The method that is best for one dataset and criterion is not the best for a different dataset and criterion. The methods that have the highest accuracy on UCSD Ped1 using the pixel-level criterion, for example, have about middle-of-the-pack accuracy on UCSD Ped2 with the pixel-level criterion.

\begin{table*}[ht]
\caption{Traditional Frame-level and Pixel-level Evaluation Criteria on the UCSD Ped1, UCSD Ped2 and CUHK Avenue Benchmark Datasets from Related Literature, Ordered Chronologically, complied from this same list. *Some of the earlier works unfortunately use only a partially annotated subset available at the time to report performance.}
\label{tab:auc-table}
\centering
\resizebox{\linewidth}{!}{%
\begin{tabular}{llllll}
\hline
\textbf{Method} & \textbf{\begin{tabular}[c]{@{}l@{}}UCSD Ped1\\ frame AUC/EER\end{tabular}} & \textbf{\begin{tabular}[c]{@{}l@{}}UCSD Ped1\\ pixel AUC*\end{tabular}} & \textbf{\begin{tabular}[c]{@{}l@{}}UCSD Ped2\\ frame AUC/EER\end{tabular}} & \textbf{\begin{tabular}[c]{@{}l@{}}UCSD Ped2\\ pixel AUC\end{tabular}} & \textbf{\begin{tabular}[c]{@{}l@{}}CUHK Avenue \\ frame AUC/EER\end{tabular}}     \\ \hline
Adam \cite{adam_robust_2008} & 65.0\%/38.0\% & 46.1\% & 63.0\%/42.0\% & 18.0\% & -/- \\ \hline
Social force \cite{mehran_abnormal_2009} & 67.5\%/31.0\% & 19.7\% & 63.0\%/42.0\% & 21.0\% & -/- \\ \hline
MPPCA \cite{mahadevan_anomaly_2010} & 59.0\%/40.0\% & 20.5\% & 77.0\%/30.0\%  & 14.0\% & -/- \\ \hline
Social force + MPPCA \cite{mahadevan_anomaly_2010} & 67.0\%/32.0\% & 21.3\% & 71.0\%/36.0\% & 21.0\% & -/- \\ \hline
MDT \cite{mahadevan_anomaly_2010} & 81.8\%/25.0\% & 44.1\% & 85.0\%/25.0\% & 44.0\% & -/- \\ \hline
Video parsing \cite{antic_video_2011} & 91.0\%/18.0\% & 83.6\% & 92.0\%/14.0\% & 76.0\% & -/- \\ \hline
Local statistical aggregates \cite{saligrama_video_2012} & 92.7\%/16.0\% & - & -/- & - & -/- \\ \hline
Detection at 150 FPS (SCL) \cite{lu_abnormal_2013} & 91.8\%/15.0\% & 63.8\% & -/- & - & -/- \\ \hline
Sparse reconstruction \cite{cong_abnormal_2013} & 86.0\%/19.0\% & 45.3\% & -/- & - & -/- \\ \hline
HMDT CRF \cite{weixin_li_anomaly_2014} & -/17.8\% & 82.7\% & -/18.5\% & - & -/- \\ \hline
AMDN \cite{xu_learning_2015} & 92.1\%/16.0\% & 67.2\% & 90.8\%/17.0\% & - & -/- \\ \hline
ST video parsing \cite{antic_spatio-temporal_2015} & 93.9\%/12.9\% & 84.2\% & 94.6\%/10.6\% & 81.1\% & -/- \\ \hline
App+motion cues \cite{zhang_combining_2016} & 85.0\%/- & 65.0\% & 90.0\%/- & - & -/- \\ \hline
Conv-AE \cite{hasan_learning_2016} & 81.0\%/27.9\% & - & 90.0\%/21.7\% & - & 70.2\%/25.1\% \\ \hline
Deep event models \cite{feng_learning_2017} & 92.5\%/15.1\% & 69.9\% & -/- & - & -/- \\ \hline
Compact feature sets \cite{leyva_video_2017} & 82.0\%/21.1\% & 57.0\% & 84.0\%/19.2\% & - & -/- \\ \hline
Conv-WTA-AE \cite{tran_anomaly_2017} & 91.9\%/15.9\% & 68.7\% & 92.8\%/11.2\% & 80.9\% & 82.1\%/24.2\% \\ \hline
RBM \cite{vu_energy-based_2017} & 70.3\%/35.4\% & 48.9\% & 86.4\%/16.5\% & 72.1\% & 78.8\%/27.2\% \\ \hline
Convex polytope ensembles \cite{battiato_convex_2017} & 78.2\%/24.0\% & 62.2\% & 80.7\%/19.0\% & 75.7\% & -/- \\ \hline
Joint detection and recounting \cite{hinami_joint_2017} & -/- & - & 92.2\%/13.9\% & 89.1\% & -/- \\ \hline
Sparse coding revisit \cite{luo_revisit_2017} & -/- & - & 92.2\%/- & - & 81.7\%/- \\ \hline
GAN \cite{ravanbakhsh_abnormal_2017} & 97.4\%/8.0\% & 70.3\% & 93.5\%/14.0\% & - & -/- \\ \hline
Online-GNG \cite{sun_online_2017}  & 93.8\%/- & 65.1\% & 94.0\%/- & - & -/- \\ \hline
Future frame prediction \cite{liu_future_2018} & 83.1\%/- & - & 95.4\%/- & - & 85.1\%/- \\ \hline
Plug and play CNN \cite{ravanbakhsh_plug-and-play_2018} & 95.7\%/8.0\% & 64.5\% & 88.4\%/18.0\% & - & -/- \\ \hline
Fast SCL \cite{lu2019fast} & 93.8\%/14.0\% & 84.1\% & 95.0\%/- & 80.0\% & -/- \\ \hline
Narrowed normality clusters\cite{ionescu_detecting_2019} & -/- & - & -/- & - & 88.9\%/- \\ \hline
Object-centric auto-encoders \cite{ionescu_object-centric_2019} & -/- & - & 97.8\%/- & - & 90.4\%/- \\ \hline
Appearance-motion cGAN \cite{nguyen_anomaly_2019} & -/- & - & 96.2\%/- & - & 86.9\%/- \\ \hline
MLAD\textsubscript{0+3} \cite{vu_robust_2019} & 82.3\%/23.5\% & 66.6\% & 99.2\%/2.5\% & 97.2\% & 71.5\%/36.4\% \\ \hline
Memory-augmented AE \cite{gong_memorizing_2019} & -/- & - & 94.1\%/- & - & 83.3\%/- \\ \hline
Prediction+reconstruction \cite{tang_integrating_2020} & 82.6\%/- & 78.4\% & 96.2\%/- & 93.1\% & 83.7\%/- \\ \hline
NN on video patch FG masks \cite{ramachandra_street_2019} & 77.3\%/25.9\% & 69.3\% & 88.3\%/18.9\% & 83.9\% & 72.0\%/33.0\% \\ \hline 
Siamese distance learning \cite{ramachandra_learning_2019} & 86.0\%/23.3\% &  80.4\% & 94.0\%/14.1\% & 93.0\% & 87.2\%/18.8\%\\ \hline
Memory-guided normality \cite{park2020learning} & -/- & - & 97.0\%/- & - & 88.5\%/- \\ \hline
\end{tabular}
}
\end{table*}

\begin{table}[ht]
\caption{Track and Region-based Area Under the ROC Curve for False Positive Rates up to 1.0 on UCSD Ped1, UCSD Ped2 and CUHK Avenue.}
\vspace{-8pt}
\label{tab:track-region-aucs}
\centering
\resizebox{\linewidth}{!}{%
\begin{tabular}{|l|lll||lll|}
  \hline
  {\bf Method} & \multicolumn{3}{c||}{\bf track AUC} & \multicolumn{3}{c|}{\bf region AUC} \\
  & {\bf Ped1} & {\bf Ped2} & {\bf Avenue} & {\bf Ped1} & {\bf Ped2} & {\bf Avenue}\\ \hline
\cite{ramachandra_street_2019} (FG masks) & 84.6\% & 80.5\% & 80.9\% & 46.6\% & 62.5\% & 35.8\% \\ \hline
\cite{ramachandra_street_2019} (Flow) & 86.5\% & 83.2\% & 78.4\% & 48.3\% & 55.0\% & 27.3\%\\ \hline
\cite{ramachandra_learning_2019} (Siamese net) & 90.0\% & 89.3\% & 78.6\% & 59.2\% & 74.0\% & 41.2\% \\ \hline
\end{tabular}%
}
\end{table}

\begin{table}[ht]
\caption{Track-based, Region-based, Pixel-level, and Frame-level Area Under the ROC Curve on Street Scene.}
\vspace{-8pt}
\label{tab:SS-aucs}
\centering
\resizebox{\linewidth}{!}{%
\begin{tabular}{|l|c|c|c|c|}
  \hline
  {\bf Method} & {\bf track AUC} & {\bf region AUC} & {\bf pixel AUC} & {\bf frame AUC} \\ \hline
\cite{hasan_learning_2016} (Autoencoder) & 2\% & 0.3\% & 0.1\% & 61\% \\ \hline
\cite{lu_abnormal_2013} (Dictionary method) & 10\% & 2\% & 7\% & 48\% \\ \hline
\cite{ramachandra_street_2019} (Flow) & 52\% & 11\% & 17\% & 51\% \\ \hline
\cite{ramachandra_street_2019} (FG masks) & 53\% & 21\% & 30\% & 61\% \\ \hline
\end{tabular}%
}
\end{table}

A combination of lack of realistic datasets and evaluation criteria has meant that progress in research in video anomaly detection has not directly translated to high-performing systems deployed in practice. Partly for this reason, reporting running times has not been common practice for research in this field. Nevertheless, Table \ref{tab:runtimes} lists running times during inference for various methods where available. Since the resolution of the frame directly affects processing time for most methods, we also list the datasets on which the runtimes are reported where available. From the table, there is a clear trend where probabilistic approaches, although they can detect anomalies in a very principled framework, struggle to perform detection in real-time.

\begin{table}[]
\centering
\caption{Running times of methods from literature, compiled from this same list.}
\label{tab:runtimes}
\resizebox{\linewidth}{!}{%
\begin{tabular}{llll}
\hline
\textbf{Method}                                       & \textbf{Approach} & \textbf{Fps} & \textbf{Dataset}                                                                                                                                             \\ \hline
\cite{sabokrou_real-time_2015}     & Dist.             & 200          & \begin{tabular}[c]{@{}l@{}}UCSD Ped1, UCSD Ped2, \\ UMN\end{tabular}                 \\ \hline
\cite{battiato_deep_2017}          & Dist.             & 20           & CUHK Avenue                                                                                                                                                  \\ \hline
\cite{sabokrou_deep-cascade:_2017} & Dist.             & 130          & UCSD Ped1, UCSD Ped2, UMN                                                                                                                                    \\ \hline
\cite{sabokrou_deep-anomaly:_2018} & Dist.             & 370          & UCSD Ped2                                                                                                                                                    \\ \hline
\cite{ionescu_detecting_2019}      & Dist.             & 24           & CUHK Avenue, Subway, UMN                                                                                                                                     \\ \hline
\cite{ionescu_object-centric_2019} & Dist.             & 11           & \begin{tabular}[c]{@{}l@{}}CUHK Avenue, UCSD Ped2, \\ ShanghaiTech, UMN\end{tabular} \\ \hline \hline
\cite{mahadevan_anomaly_2010}      & Prob.             & 0.4          & UCSD Ped2                                                                                                                                                    \\ \hline
\cite{antic_video_2011}            & Prob.             & 0.13         & UCSD Ped1                                                                                                                                                    \\ \hline
\cite{weixin_li_anomaly_2014}     & Prob.             & 1.25         & UCSD Ped2                                                                                                                                                    \\ \hline
\cite{antic_spatio-temporal_2015}  & Prob.             & 1            & UCSD Ped1, UCSD Ped2                                                                                                                                         \\ \hline
\cite{cheng_video_2015}            & Prob.             & 2            & UCSD Ped1                                                                                                                                                    \\ \hline \hline
\cite{chong_abnormal_2017}         & Recon.            & 143          & \begin{tabular}[c]{@{}l@{}}CUHK Avenue, Subway, \\ UCSD Ped1, UCSD Ped2\end{tabular} \\ \hline
\cite{liu_future_2018}             & Recon.            & 25           & CUHK Avenue                                                                                                                                                  \\ \hline
\cite{lu_abnormal_2013}            & S-Recon.          & 150          & CUHK Avenue                                                                                                                                                  \\ \hline
\cite{cong_abnormal_2013}          & S-Recon.          & 0.26         & UCSD Ped1                                                                                                                                                    \\ \hline
\cite{luo_revisit_2017}            & S-Recon.          & 50           & UCSD Ped2                                                                                                                                                    \\ \hline
\cite{gong_memorizing_2019}        & S-Recon.          & 38           & UCSD Ped2                                                                                                                                                    \\ \hline
\cite{lu2019fast}                    & S-Recon.          & 1000         & \begin{tabular}[c]{@{}l@{}}UCSD Ped2, CUHK Avenue, \\ Subway\end{tabular}            \\ \hline
\end{tabular}%
}
\end{table}

\section{Discussion}
\label{sect:discussion}
We have provided a comprehensive review of research in single-view video anomaly detection. We built an intuitive taxonomy and situated past research works in relation to each other. We also hope this article will serve to clear up some misconceptions among different problem formulations, use of datasets, evaluation protocol and how to compare against methods that use compatible problem formulation and evaluation schema in their assumptions. We now provide some best practices and state some observations on the evolution of the field in terms of overarching trends in representation and modeling as they relate to the increasing size of datasets and increasing compute power of devices.

\subsection{Best Practices Going Forward}
In terms of future best practices, we urge researchers in this area to use the recommended reliable datasets, new evaluation protocol and participate in reproducible research. As the field matures into producing approaches that are viable in practice, researchers should also provide runtime analyses of their methods. A qualitative evaluation of quality of false positives is also important, especially to discover biases in modeling.  Evaluating on multiple datasets is essential; for example, some works that evaluate solely on UCSD Ped1, UCSD Ped2 and UMN datasets are known to be inherently biased towards the anomalies in these datasets, which are mainly comprised of objects with larger motion magnitudes. CUHK Avenue and Street Scene have emerged as good supplements with more variation in anomalous activity.

\subsection{Trends in Representation}
Representation of input to video anomaly detection algorithms was mostly dominated by raw, fixed-size image patches. Some anomalies require analyzing temporal information, so researchers turned to using video patches, which required more compute power. More recently, researchers have started using multi-modal representations of video patches, with raw frames as well as estimated optical flow fields to the point where it is the norm now. Some methods have even attempted to use entire frames and video snippets as input by exploiting advances in GPU compute power. We expect this trend of the increasing complexity of input representation to reverse with the use of 3D and inflated 3D convolutions on raw video (foregoing expensive optical flow field computation) which have become popular in video action recognition \cite{carreira2017quo}.

\subsection{Trends in Modeling}
Meanwhile, modeling has followed a different trend. At first, researchers used very simple hand-crafted features whose distribution could be well modeled with simple assumptions. Soon researchers achieved better results with more complex models, more intricate assumptions and a lot of clever engineering. More recently, the trend has reversed, with a larger reliance on learning representations from data to more directly optimize a cleverly set up optimization scheme and elegant modeling approach. We expect this trend of having the data dominate to continue, especially as larger, more complex datasets become available. 

\subsection{Looking Ahead}
On one hand, video anomaly detection research has come a long way. On the other hand, past research has also neglected tackling some of the more challenging problems in video anomaly detection. In existing datasets, loitering anomalies have not exactly been addressed in specific by modeling. In fact, most past approaches are unable to detect these kinds of anomalies since they rely heavily on motion cues to ignore processing parts of the video. Working on an algorithm to retain the benefits of any recent state of the art method that is also able to detect loitering anomalies is one ripe area for contribution. Of note is one recent work by Rodrigues et al. \cite{rodrigues2020multi} that has attempted to address loitering anomalies in specific by modeling activities in a multi-timescale manner. Another challenge for video anomaly detection methods is the ability to handle rare but normal activity. Such activity, which may appear very sparsely in the normal training video, often causes false positive anomaly detections.  An example of such activity is a pedestrian stopping to tie her shoe.  This probably does not happen very often and a security guard may not want the anomaly detector to raise an alarm when it does. So the model that is learned from normal video should include not only the most common normal activities but rare, normal activities as well.

In terms of the types of anomalies, group, trajectory and time of day anomalies have largely been unaddressed because benchmark datasets that contain these simply do not exist yet. We urge and expect other researchers to contribute datasets with these properties in the near future.

As researchers move on from a focus on smaller, less complex datasets for which accuracy is becoming saturated, to larger, more complex datasets with a greater variety of anomaly types, they will be pushed to invent new video representations and new modeling strategies that can achieve high detection rates at low false positive rates to make algorithms that are practical for real applications.

{\small
\bibliographystyle{IEEEtran}
\bibliography{Survey,extras}
}

%

\begin{IEEEbiography}[{\includegraphics[width=1in,height=1.25in,clip,keepaspectratio]{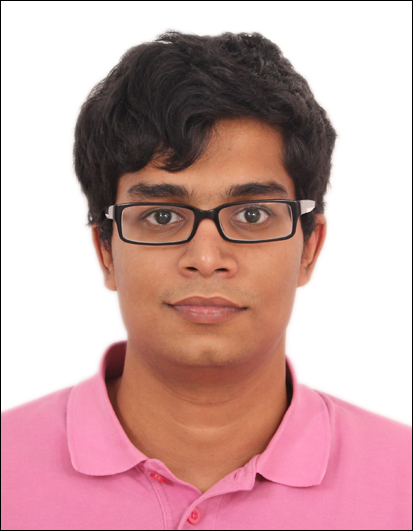}}]{Bharathkumar Ramachandra}
Bharathkumar ``Tiny'' Ramachandra received his Ph.D. in 2020 from North Carolina State University. He is currently a computer vision scientist at Wrnch AI in Montreal, Quebec.  His research interests are broadly machine learning and computer vision and in particular video anomaly detection, metric learning, deep generative modeling and out-of-distribution generalization with neural networks.
\end{IEEEbiography}

\begin{IEEEbiography}[{\includegraphics[width=1in,height=1.25in,clip,keepaspectratio]{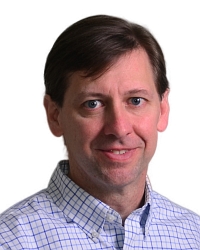}}]{Michael Jones}
Michael Jones received his Ph.D. in 1997 from the Massachusetts Institute of Technology.  He was a member of DEC's Cambridge Research Lab from 1997 to 2001 and is currently a senior principal research scientist at Mitsubishi Electric Research Laboratories (MERL) in Cambridge, Massachusetts where he has been since 2001.  He has published papers in many areas of computer vision and machine learning and is mainly interested in object detection, action detection, face recognition, person re-identification and video anomaly detection.
\end{IEEEbiography}


\begin{IEEEbiography}[{\includegraphics[width=1in,height=1.25in,clip,keepaspectratio]{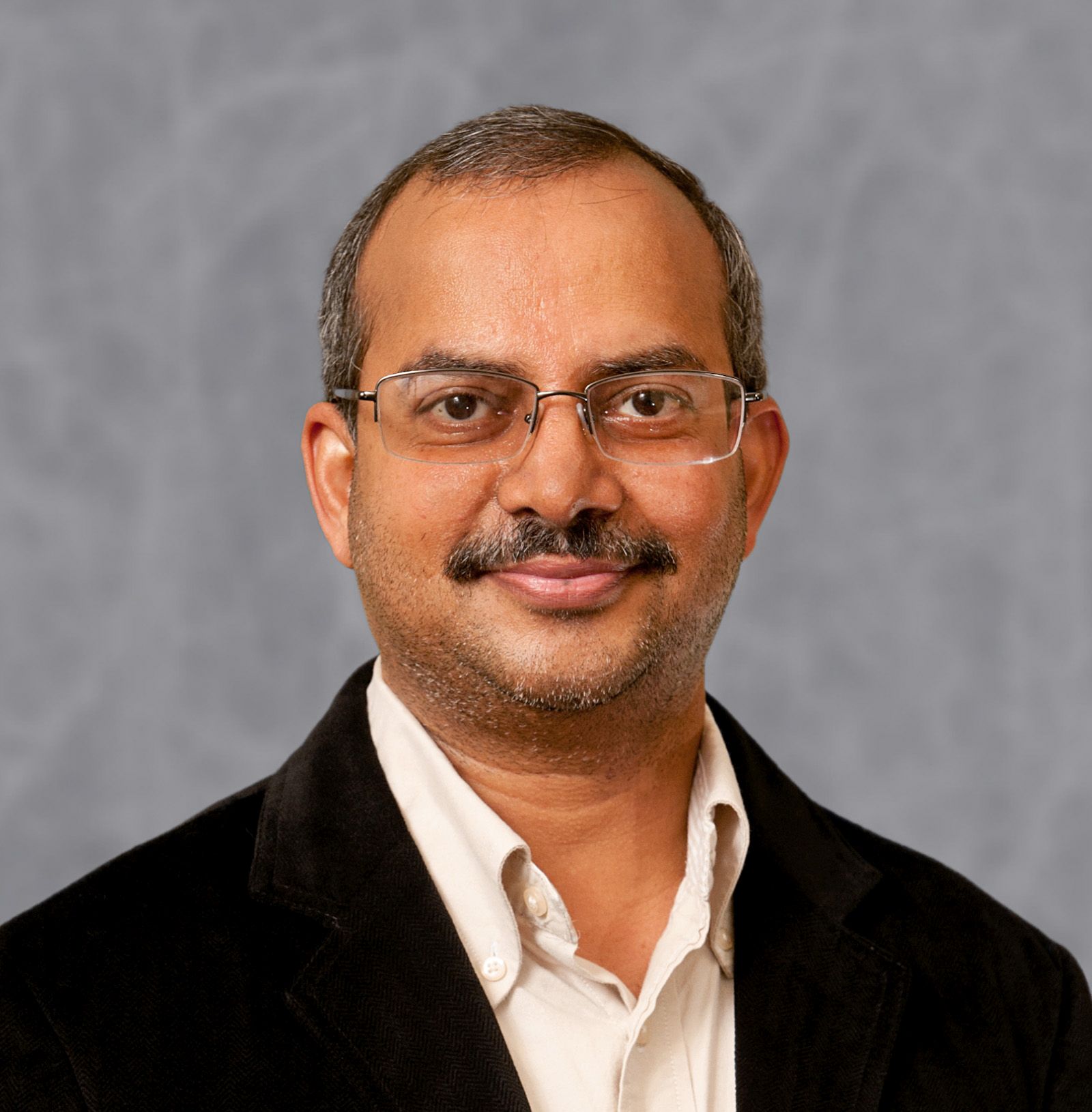}}]{Ranga Raju Vatsavai}
Raju is a Chancellor’s Faculty Excellence Program Cluster Associate Professor of Geospatial Analytics in the Department of Computer Science, North Carolina State University (NCSU). Before joining NCSU, Raju was the Lead Data Scientist for the Computational Sciences and Engineering Division (CSED) at the Oak Ridge National Laboratory (ORNL). He works at the intersection of spatial and temporal big data management, machine learning, and high-performance computing with applications in national security, geospatial intelligence, natural resources, agriculture, climate change, location-based services, and human terrain mapping. He holds MS and PhD degrees in computer science from the University of Minnesota.
\end{IEEEbiography}

\vfill




\end{document}